\documentclass[acmlarge,nonacm]{acmart}

\usepackage{algorithm}
\usepackage{algorithmic}

\usepackage{enumitem}
\usepackage{makecell}
\usepackage{multirow}
\usepackage{caption}
\usepackage{subcaption}
\usepackage{colortbl}
\usepackage[]{xcolor}
\usepackage{xcolor}
\usepackage[normalem]{ulem} 

\AtBeginDocument{%
  \RequirePackage{hyperxmp}
}
\begin{document}

\title{Human Heterogeneity Invariant Stress Sensing}


\author{\href{https://orcid.org/0000-0002-5261-5440}{Yi Xiao}}
\email{yxiao54@syr.edu}
\affiliation{%
  \institution{Arizona State University}
  \city{Tempe}
  \state{Arizona}
  \country{USA}
  }

\author{\href{https://orcid.org/0000-0002-7016-6220}{Harshit Sharma}}
\email{hsharm04@syr.edu}
\affiliation{%
  \institution{Arizona State University}
  \city{Tempe}
  \state{Arizona}
  \country{USA}
  }
\author{\href{https://orcid.org/0009-0002-8937-3702}{Sawinder Kaur}}
\email{sakaur@syr.edu}
\affiliation{%
  \institution{Syracuse University}
  \city{Syracuse}
  \state{New York}
  \country{USA}
  }

\author{\href{https://orcid.org/0000-0002-8852-732X}{Dessa Bergen-Cico}}
\email{dkbergen@syr.edu}
\affiliation{%
  \institution{Syracuse University}
  \city{Syracuse}
  \state{New York}
  \country{USA}
  }
  
\author{\href{https://orcid.org/0000-0002-0807-8967}{Asif Salekin}}
\thanks{Asif Salekin is the corresponding author}
\email{asalekin@syr.edu}
\affiliation{%
  \institution{Arizona State University}
  \city{Tempe}
  \state{Arizona}
  \country{USA}
  }


\begin{abstract}

Stress affects physical and mental health, and wearable devices have been widely used to detect daily stress through physiological signals. \textcolor{black}{However, these signals vary due to factors such as individual differences and health conditions, making generalizing machine learning models difficult. To address these challenges, we present Human Heterogeneity Invariant Stress Sensing (HHISS), a domain generalization approach designed to find consistent patterns in stress signals by removing person-specific differences. This helps the model perform more accurately across new people, environments, and stress types not seen during training. Its novelty lies in proposing a novel technique called person-wise sub-network pruning intersection to focus on shared features across individuals, alongside preventing overfitting by leveraging continuous labels while training. The study focuses especially on people with opioid use disorder (OUD)—a group where stress responses can change dramatically depending on their time of daily medication taking. Since stress often triggers cravings, a model that can adapt well to these changes could support better OUD rehabilitation and recovery. We tested HHISS on seven different stress datasets—four of which we collected ourselves and three public ones. Four are from lab setups, one from a controlled real-world setting, driving, and two are from real-world in-the-wild field datasets without any constraints. This is the first study to evaluate how well a stress detection model works across such a wide range of data. Results show HHISS consistently outperformed state-of-the-art baseline methods, proving both effective and practical for real-world use. Ablation studies, empirical justifications, and runtime evaluations confirm HHISS's feasibility and scalability for mobile stress sensing in sensitive real-world applications.}

\end{abstract}


\maketitle

\section{Introduction}
\label{introduction}
Stress, while sometimes beneficial, can lead to serious mental and physical health issues when prolonged or unmitigated \cite{mcewen2008central}. With the rise of wearable technology, there is growing interest in using smartwatch-type devices to monitor psychophysiological responses, such as electrodermal activity (EDA), heart rate, and skin temperature, linked to the human body’s stress response \cite{buijs2000integration,sonkusare2019detecting,mauri2010psychophysiological}. Continuous monitoring of these signals, combined with machine learning, offers a promising path for automated stress detection \cite{campanella2023method,gedam2021review}.
\par 
However, deploying stress detection models in real-world settings is challenging due to the inherent heterogeneity in stress responses, which differ both between and within individuals \cite{lecic2011stress,gillespie2009risk,jacoby2021individual,nagaraj2023dissecting}. 
Factors such as age, health, medication, stressor type, and environmental conditions, like temperature and noise, lead to distribution or domain shifts in wearable stress sensing data \cite{foltyn2021towards,saeed2018model,yu2022semi}, further complicating the detection \cite{chandra2021comparative,mccarthy2016validation}.
Personalization approaches, which tailor models to individual users, may address these distribution shifts \cite{bolpagni2024personalized,tazarv2021personalized}. While these methods can enhance model performance, they are often constrained by the availability of user-specific data, which are scarce and often unavailable for sensitive health populations. Additionally, such approaches may not generalize well in the face of intra-individual variability or distribution shifts, influenced by  stressors and changes in medication use \cite{nagaraj2023dissecting,chapman2022impact}.
\par 
This challenge is further exacerbated in individuals with opioid use disorder (OUD), one of the most pressing contemporary public health challenges. In 2022, there were over 80,000 opioid-related overdose deaths in the United States, averaging more than 224 each day \cite{nguyen2024trends}. There were an estimated $6.1$ million people with an OUD in the U.S. in $2022$ \cite{samhsa2023key}. Higher stress is a known predictor of opioid craving \cite{maclean2019stress}; hence, automated monitoring of OUD individuals' stress can prevent them from relapsing and significantly improve their rehabilitation.
However, the use of opioids produces long-term and acute effects that alter physiological patterns \cite{poh2012convulsive,wang2023cardiac}, causing the exhibition of greater physiological stress responses compared to the healthy controls without an OUD \cite{back2015laboratory}. Additionally, stress responses are heterogeneous within the OUD group \cite{maclean2019stress}. This variability, influenced by factors like altered stress regulation and co-occurring health conditions, deviates significantly from typical patterns seen in non-OUD individuals \cite{carlson2012trait,davis2017prescription}. Particularly, the medication for opioid use disorder (MOUD) adds layers of complexity to stress detection in this population \cite{national2018medication,maclean2019stress}. For example, the same individuals with OUD may exhibit altered stress responses due to various factors, especially in relation to their daily MOUD doses, resulting in intra-individual variability before and after medication intake \cite{carreiro2016wearable,chintha2018wearable}. Given the variabilities, machine learning models trained on general population data may not effectively predict stress in OUD individuals, reducing their clinical utility. Moreover, collecting comprehensive stress response data across all possible situations of OUD individuals for personalization is impractical, especially due to ethical, logistical, and recruitment barriers \cite{avery2024barriers}. The lack of diverse datasets comprising OUD individuals further hinders model development,

Domain generalization (DG), also known as out-of-distribution (OOD) generalization \cite{jiang2022transferability,tllib,wang2022provable} offers a potential solution by training models to perform well across diverse populations, even when trained only on general healthy (i.e., control) population data \cite{wang2022generalizing}. 
While generic OOD robustness approaches in existing literature have been applied to human-centered data to extract distribution-invariant features \cite{sanner2021reliable,varnava2025out,yang2022multimodal}, they often fail to account for the heterogeneity across individuals. For instance, a stress model developed using data from a generally healthy population may overlook the physiological characteristics of individuals who experience higher sweating since they are underrepresented in the training set. However, higher sweating is an impact of opioid medication \cite{pergolizzi2020opioid}, and by neglecting these underrepresented characteristics during feature extraction, the model becomes less generalizable to individuals with Opioid Use Disorder (OUD) following their dose of medication for opioid use disorder (MOUD).

To address the heterogeneity challenge in human-centered stress sensing with DG or OOD generalization, this paper introduces \emph{Human Heterogeneity Invariant Stress Sensing (HHISS)}, which uniquely addresses how individual differences impact the extraction of invariant stress features, avoids overfitting on characteristics predominant in the majority and offers a novel solution that captures \emph{equitable invariant} features across all. Specifically, \emph{the HHISS approach is trained exclusively on a healthy population exposed to specific stressors (e.g., lab-induced stimuli). Despite this limited training, the model generalizes effectively across changes in stressors, environments, and individuals not represented in the training data, including those experiencing real-world stress due to other stressor like driving and the stress responses of individuals with OUD, both before and after MOUD intake, among others.}

This generalizability represents a step forward in creating inclusive and reliable stress detection systems. By demonstrating OOD generalizability across diverse OUD conditions, HHISS extends the applicability of OOD stress sensing models trained on healthy individuals to health conditions or populations for which data may be scarce, thereby enhancing the accessibility and effectiveness of stress management tools across a wider range of contexts. Notably, just like other DG or OOD approaches, HHISS complements personalization techniques by providing a strong baseline that performs well even \textcolor{black}{with only a few user samples.}

The main contributions and empirical findings of the present study are:
\begin{enumerate}

\item This work presents \emph{HHISS}, a novel framework that addresses the inherent heterogeneity in psychophysiological stress responses. By learning \emph{equitable} invariant features through a unique intersectional person-wise sub-network extraction alongside contemporary optimization paradigms in a unique way, HHISS achieves superior OOD stress sensing robustness (Section \ref{our-approach}).

\item This paper examines the OOD robustness of HHISS in the OUD population under varying MOUD conditions, showcasing its role in improving stress sensing accessibility and effectiveness. To facilitate the investigation, we collected data from 25 OUD participants across 29 sessions (pre- and post-medication) and 46 healthy controls, forming three distinct lab-based distributions. Our findings show that extensive disease-condition data is not essential for developing robust OOD stress detection models, as collecting such data is often impractical. Notably, training HHISS on only pre- or post-medication data resulted in suboptimal performance, while leveraging diverse control data for training significantly improved accuracy across distributions (Section \ref{RQ-2-other-datasets-training}). HHISS’s human-centered design ensures OOD robustness using only control data for training, making it highly valuable for health applications with limited data availability.



\item We further evaluated HHISS on two public datasets with diverse stressors, environments, and populations, demonstrating its OOD robustness beyond training conditions (Section \ref{rq3}), and its ability to learn equitable invariant features is not limited to one dataset (Section \ref{rq4}), highlighting its versatility and reliability. 

\item \textcolor{black}{In addition, we evaluated HHISS on two real-world in-the-wild field datasets without any constraints. One publicly available nurse work stress dataset, and one where we collected five days of continuous 24-hour data from two individuals with OUD, enabling assessment of intra-individual generalizability across different times of day. HHISS achieves superior performance on both datasets, demonstrating superior OOD robustness and generalizability under complex, real-world conditions, including for OUD individuals over time of the day. To our knowledge, this is the first paper to assess OOD robustness across such a comprehensive set of human-stress-sensing data distributions.}

\item Comparison with state-of-the-art OOD robustness and unsupervised domain adaptation baselines demonstrates that HHISS strikes a superior balance, excelling in both in- and out-of-distribution scenarios (Section \ref{results-baselines}). Additionally, our ablation study (Section \ref{results-ablation}), runtime, and training-time analysis (Section \ref{table:training_platform_comparison}) confirm HHISS's feasibility and scalability, highlighting its potential to enable mobile stress sensing in real-world, sensitive applications where data access is limited. 

\item We empirically validated (Section \ref{results-empirical-justification}) the assumptions that led to HHISS's design choices outlined in Section \ref{our-approach} and conducted post-analysis (Section \ref{distribution-shif-validation}) to confirm the OOD challenges in generic stress sensing and demonstrate HHISS's ability to mitigate it.

\end{enumerate}

\section{Related-works}\label{related-work}
 \subsection{Wearable sensor-based stress sensing}\label{Wearable sensor-based stress sensing}

Smart wearables have been extensively studied for stress detection \cite{ollander2016comparison,menghini2019stressing}, using sensors like electrodermal activity (EDA), heart rate, and skin temperature to provide real-time, objective data. These insights are valuable for clinicians, complementing behavioral interventions \cite{haleem2021biosensors}. The variability in stress responses underscores the need for robust approaches that account for diverse psychophysiological characteristics.

\subsubsection{Heterogeneity in Psychophysiological Stress Response}

Stress responses vary both within individuals \cite{jacoby2021individual,sharma2022psychophysiological} and across populations \cite{sharma2022psychophysiological,perez2022gender,crielaard2021understanding,haigis2010aging}, influenced by various factors and stress-inducing tasks or stressors \cite{giles2014stress}. Individual differences in physiological stress responses are shaped by genetics \cite{gillespie2009risk}, personality traits like neuroticism and resilience \cite{jacoby2021individual,weber2022physiological,schwerdtfeger2019episodes}, psychosocial factors such as social support and coping mechanisms \cite{weber2022physiological,passarelli2021responses}, as well as environmental influences \cite{jacoby2021individual} and health status \cite{weber2022physiological,jacoby2021individual}. At the population level, stress response variability or distribution shifts are driven by gender \cite{perez2022gender,passarelli2021responses}, age \cite{haigis2010aging}, and socioeconomic status \cite{crielaard2021understanding}. Additionally, the choice of stress induction paradigms, such as the Trier Social Stress Test (TSST) \cite{allen2017trier}, can lead to varied physiological and psychological responses, with different stressors eliciting distinct magnitudes and durations of stress reactions based on individual characteristics \cite{giles2014stress}.

\subsection{Heterogeneity in Neurophysiology of OUD Individuals}\label{realted-work-OUD}

Opioids, as central nervous system (CNS) depressants, reduce heart rate, respiration, muscle tension, and blood pressure while causing peripheral vasodilation, lowered body temperature, and skin flushing. Medication-assisted treatments (MATs) like methadone are prescribed at controlled doses to prevent withdrawal symptoms with minimal psychoactive effects. Administered orally, MATs take 30-45 minutes to act and are long-lasting agonists \cite{rosenthal2022drugs}, potentially leading to blunted stress responses in individuals who completed stress tasks shortly after receiving their daily dose. In this paper, we refer to such individuals as post-dose opioid use disorder (post-dose OUD) participants.

Research has examined the physiological changes linked to drug intake, with some studies focusing on machine-learning algorithms to detect drug use and withdrawal events. 
For example, \citet{chapman2022impact} used the Empatica E4 to monitor opioid patients, revealing significant pre- and post-administration changes such as increased accelerometer frequency, skin temperature, EDA variation, and decreased heart rate. \citet{carreiro2016wearable} found reduced movement in frequent opioid users, while \citet{carreiro2015imstrong} observed increased EDA, locomotion, and lower skin temperature, following cocaine use. \textcolor{black}{These physiological fluctuations pose significant challenges for stress detection models based on physiological signals in OUD populations.}

\subsection{Domain generalization}\label{rel-work-DG}
Domain generalization (DG), also known as out-of-distribution (OOD) generalization \cite{jiang2022transferability,tllib,wang2022provable}, aims to overcome cross-distribution variability and ensure that models can generalize to unseen (OOD) domains after being trained on a limited number of training domains \cite{xu2023globem,suh2022adversarial,lu2022semantic,akrout2023domain,wang2023surban,wu2022multi}. This is often achieved by learning invariant features that remain consistent across different domains. Invariant Risk Minimization (IRM) \cite{arjovsky2019invariant}, a well-established method for domain generalization, encourages models to learn features that lead to optimal and consistent performance across diverse domain by focusing on invariant features. Risk Extrapolation (REx) \cite{krueger2021out}, a follow-up work of IRM, modifies the penalty term in the loss function to further improve generalization. Group Robust Optimization (GroupDRO) \cite{sagawa2019distributionally} focuses on improving worst-case performance by tracking domain-specific losses during training. \textcolor{black}{Another recent method, Clustering \cite{xu2023globem}, uses a convolutional autoencoder (DCEC) to group similar user data into distinct clusters. Each cluster is then processed by a dedicated Siamese network \cite{koch2015siamese} trained specifically for classification within that group. \citet{li2018learning} introduces MLDG (Meta-Learning for Domain Generalization), a method that applies a meta-learning strategy to address domain generalization. The approach simulates domain shift by splitting the training domain data into meta-train and meta-test sets, enabling the model to learn domain-invariant features.}

Pruning has been shown to improve generalization \cite{cheng2024survey,liang2021pruning}, with methods like Progressive Gradient Pruning (PGP) \cite{granger2021progressive} demonstrating strong results in domain generalization. Combining sparse training with IRM has also proven effective in preventing overfitting and enhancing generalization \cite{zhou2022sparse}. Additionally, knowledge distillation has been explored to further boost model robustness \cite{reddy2024towards,belal2021knowledge,ruder2017knowledge}.
\par 
\emph{Our evaluation considers the above-mentioned approaches as OOD robustness baselines.}


\subsection{Domain Generalization vs Adaptation}\label{generlization_Vs_adaptation}
Similar to DG, domain adaptation (DA) handles domain variability but differs in approach \cite{lu2022semantic,akrout2023domain}. It 
optimizes model performance on a specific target domain by leveraging both source and limited target data, employing feature-, loss-, and data-centric methods to reduce the source-target gap \cite{hoffman2014asymmetric,ramponi2020neural}, which is out of the scope of this paper.
However, recent unsupervised DA methods, like Test-Time Personalization (TTP) \cite{bao2024adaptive,xu2023joint}, adapt to target distributions during inference (not during training) using unannotated target data, making them ideal for privacy-sensitive health applications. \emph{This paper includes TTP as a baseline.}

\textcolor{black}{\citet{meegahapola2024m3bat} recently proposed M3BAT, an unsupervised domain adaptation method for multi-modal human sensing using Multi-Branch Adversarial Training. Unlike TTP \cite{xu2023joint}, which adapts to target domain distributions during inference, M3BAT requires access to unannotated target data during training. In contrast, our work focuses on domain generalization, operating without any access to target domain data at training time. Due to this fundamental difference in problem setting, M3BAT is not a suitable baseline for our work.}

\textcolor{black}{Instead, we adopt the Domain-Adversarial Neural Network (DANN) \cite{ganin2016domain}, which serves as the backbone of M3BAT, as one of our baselines. DANN performs adversarial training by jointly optimizing a domain discriminator and a feature extractor: the discriminator learns to distinguish between domains, while the feature extractor is trained to produce domain-invariant representations that confuse the discriminator. In our setting, each subject is treated as a separate domain, encouraging the model to extract features that generalize across individuals. Importantly, DANN does not require data from the target domain during training, making it a suitable baseline for our task.}

While both DG and DA aim to bridge the performance gap caused by domain shifts, DG is more challenging due to its stricter constraints—primarily, the absence of target domain data during model training \cite{wang2022generalizing,niemeijer2022domain}. In this work, we are \textit{ performing DG to develop a generalizable stress detection model trained \underline{only} using healthy control population, which can work well in unseen distribution shifts, including for health populations like people with OUD}. It is important to note that the response to affective stimuli within health populations like people with OUD is not only different from healthy populations \cite{wang2010alterations} but shows physiological variations with respect to their treatment, even for the same individual / subject \cite{mccaul1982short}, making it a challenging task to build inclusive machine learning models for a diverse population.

\section{Description of the Datasets} \label{Dataset-description}

To evaluate the stress detection model’s performance across both healthy individuals and a diverse range of Opioid Use Disorder (OUD) individual states/conditions, we collected data from three distinct populations: (1) healthy control individuals, (2) Pre-dose OUD participants, and (3) Post-dose OUD participants. To further assess the robustness of the model, we included two publicly available off-the-shelf datasets: the WESAD dataset and the AffectiveROAD dataset. Data collection across all datasets was performed using the Empatica E4 wristband\footnote{\url{https://www.empatica.com/research/e4/}}\cite{mccarthy2016validation}, which is equipped with an Electrodermal Activity (EDA) sensor, a photoplethysmography (PPG) sensor, an accelerometer, and a skin temperature sensor.

\subsection{Original Data Collection Datasets}\label{self-dataset-discussion}

Our self-collected three datasets followed a structured data collection protocol as outlined below.

\subsubsection{Data Collection Protocol} Following prior work \cite{xiao2024reading,zhao2023affective,almazrouei2023method}, we designed a stress stimulus protocol incorporating various stressors to capture psychophysiological responses. The tasks began with watching a calming video \cite{xiao2024reading}, performing a simple counting task \cite{sharma2022psychophysiological,tumanova2019autonomic}, and answering basic questions such as name, weather, and breakfast \cite{almazrouei2023method}. These were categorized as \emph{Non-stress tasks}. The subsequent tasks were presented in the following order: viewing a passive stress-inducing video \cite{schaefer2010assessing,xiao2024reading}, completing Stroop tasks \cite{tulen1989characterization}, the Trier Social Stress Test (TSST) \cite{kirschbaum1993trier}, and recalling a distressing memory \cite{connolly2018negative,kross2012asking}. These were categorized as \emph{Stress-inducing tasks}. These tasks encompass cognitive load, social and emotional stressors, representing a broad spectrum of everyday stress experiences. Participants' psychophysiological responses during non-stress and stress-inducing tasks were annotated as non-stress and stress signals, respectively.

\subsubsection{Participant Categories}\label{participant-description-OUD} The self-collected datasets included participants from both healthy populations and those with opioid use disorder (OUD). The participants were categorized into three groups, representing three datasets: a healthy control group and two OUD groups classified as pre-dose and post-dose, based on whether data collection occurred before or after the administration of Medication-Assisted Treatment (MAT).

\subsubsection{Participants Description} We recruited 25 participants with opioid use disorder (OUD) for a total of 29 sessions, with four participants contributing to both Pre- and Post-dose sessions. The Pre-dose group comprised 14 sessions, while the Post-dose group consisted of 15 sessions. The participants were drawn from a MOUD (Medication for Opioid Use Disorder) program \cite{leshner2019medication}, where they received a daily dose of methadone as part of their treatment plan. Since the neurophysiological responses to stress in these participants could vary depending on the timing of their last dose \cite{carreiro2016wearable,fishman2000adherence}, pre- and post-dose sessions were treated as two separate datasets, representing distinct distributions of stress responses in OUD populations. \textcolor{black}{For participants who took part in both the Pre-dose and Post-dose sessions, there is a 1 to 3-month gap between recordings. Due to this extended interval and the potential for day-to-day variability in physiological patterns, we treat each visit as an independent session.}  Additionally, we recruited 46 healthy participants for 46 sessions. Data collection was conducted indoors, with all participants wearing the Empatica E4 wristband on both wrists throughout the experiments. The tasks were designed in collaboration with an addiction specialist and a behavioral psychologist and were approved by the X University Institutional Review Board (IRB) for ethical compliance.

\subsection{Publicly Available Datasets}\label{dataset:public}
To further assess the broader impact of our research, we included evaluations on the WESAD and AffectiveRoad datasets, even though the primary focus is on the OUD dataset. 
These comparisons evaluate the model's ability to generalize across different datasets, adapting to unseen stressors and environments.


\subsubsection{WESAD Dataset} \cite{schmidt2018introducing} This dataset, designed for wearable stress and affect detection, includes data from 15 subjects who participated in baseline, amusement, and stress tasks as part of a stress elicitation protocol. For this study, we focused on the baseline and stress tasks to perform binary classification. The baseline task involved the participants sitting or standing while reading neutral material, and the stress task included public speaking and mental arithmetic, both known stress-inducing activities.

\subsubsection{AffectiveROAD Dataset} \cite{haouij2018affectiveroad}: Collected in Tunisia, this dataset comprises 13 driving sessions from 10 drivers. The data collection involved a 30-minute parking period (considered as a non-stress task) followed by city driving. While driving, participants’ stress annotations were constructed in real-time by an observer seated in the rear seat of the car. Annotated data is considered as the stress task signal.

\subsection{\textcolor{black}{In-the-wild field dataset}}
\textcolor{black}{To further assess the generalizability of our approach in real-world in-the-wild field conditions, we further evaluate our model on one publicly available in-the-wild field stress dataset and a self-collected OUD in-the-wild field dataset.}

\subsubsection{\textcolor{black}{OUD Daily stress dataset}}\label{OUD-field-dataset}

\begin{figure}[h!]
    \centering
    \includegraphics[scale=0.38]{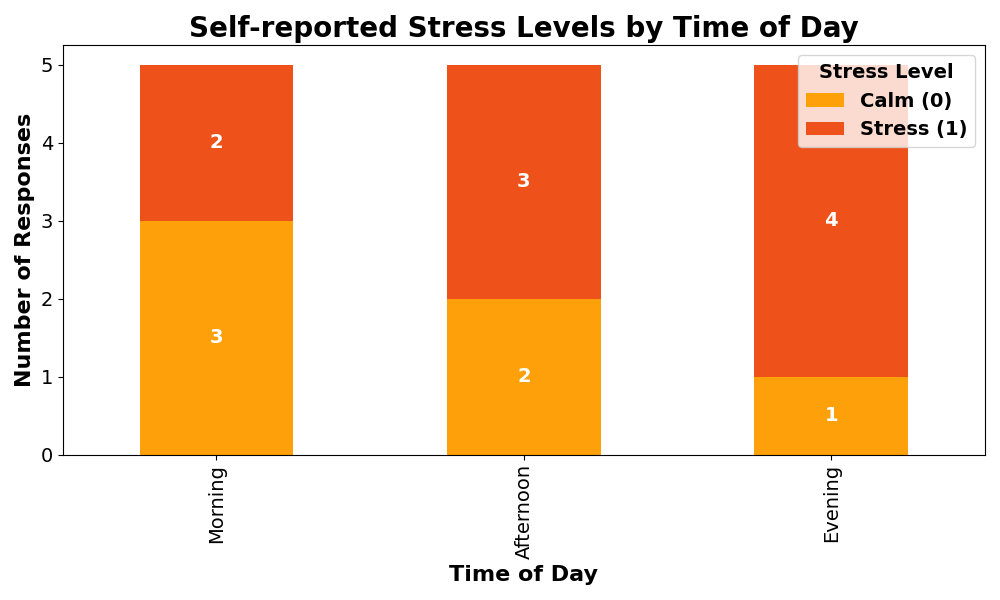}
    \caption{\textcolor{black}{Distribution of self-reported stress labels across different times of day during in-the-wild field data collection for two OUD participants of the OUD Daily stress dataset.}}
    \label{fig:ema}
    \vskip -3ex
\end{figure}

\textcolor{black}{We recruited two new unique participants with OUD to wear the Empatica EmbracePlus device (an updated version of the Empatica E4) continuously throughout the day over two and three days, respectively. The data collection was approved by the X University Institutional Review Board (IRB) for ethical compliance, and recruitment followed the same criteria as in the lab-based experiments. To capture naturally occurring stress, participants received Ecological Momentary Assessment (EMA) prompts three times daily—at 8 a.m., 1 p.m., and 6 p.m.—asking them to report whether they experienced stress in the previous two hours (0 = no stress, 1 = stressed). No restrictions were imposed during the study period to allow for spontaneous, real-world data collection. Figure \ref{fig:ema} shows the cumulative distribution of EMA responses from both participants grouped by time of day.
}

\subsubsection{\textcolor{black}{Nurses work Stress Dataset}}\label{nurse-dataset} \cite{hosseini2022multimodal}: \textcolor{black}{This dataset investigates stress in a workplace setting influenced by various social, cultural, and psychological factors. It contains 1,250 hours of physiological data collected from 15 nurses. After each shift, nurses completed a survey reporting their stress levels along with the start and end times of their shift. For this study, we used only the labeled data recorded during the shifts, resulting in approximately 172 hours of usable data. The remaining data was unlabeled and excluded from our analysis.}

\subsection{Datasets' Distribution Variability}

These five datasets incorporate a broad range of factors that can influence stress responses' variability, thereby representing distinct distributions of stress. Such variability is critical in assessing the performance of stress detection models across different distinct distributions. For instance, the pre-dose and post-dose datasets differ due to the effects of drug usage, while the WESAD dataset exhibits variability arising from different laboratory setups and institutional environments. The AffectiveROAD dataset presents real-world conditions characterized by complex environments and unique stressors.
Our analyses in sections \ref{motivation} validates the presence of stress response distribution (and signal features) variability/shifts across these datasets.

\subsection{Feature extraction}
\label{sec:feature}


\textbf{ Preprocessing }: The Empatica E4 wristband collects data from four modalities: Electrodermal Activity (EDA), Photoplethysmography (PPG), an accelerometer, and a skin temperature. PPG and EDA data were cleaned and processed for accurate feature extraction. A Butterworth filter (0.5-8 Hz) was applied to PPG signals, with gaps filled using quartic interpolation \cite{campanella2023method,wang2023detecting}. For EDA, a 4th-order Butterworth low-pass filter (3 Hz) was applied \cite{campanella2023method}, and skin conductance response onset and peaks were detected using a 0.05 Siemens threshold to ensure precise stress-related physiological responses identification \cite{makowski2021neurokit2}.

\textbf{Feature Extraction}: In the field of physiological signal-based human sensing, larger window sizes can produce more reliable features \cite{campanella2023method,flirt2021,makowski2021neurokit2}, while smaller window sizes enable real-time stress detection. Considering the experimental protocol, we chose a 30-second window with a 15-second overlap for stress detection.

Following previous stress detection approaches \cite{beierle2023automating,flirt2021,strvzinar2023stress,han2024systematic}, we extract 340 statistical features (after removal of some features inappropriate to extract from a 30-second window, such as HRV SDANN5, which requires at least 5 minutes of data \cite{makowski2021neurokit2}, and non-linear features that require a minimum of 2 minutes of data \cite{bernardes2022reliable}. 
These features capture key physiological indicators of stress and anxiety, which can vary significantly between individuals \cite{boucsein2012electrodermal}. To ensure comparability, we normalize data for each subject. To mimic real-world scenarios where full datasets aren't available, we apply change score normalization \cite{sharma2022psychophysiological}. A baseline is established using each subject's first three windows from a calm session, which are excluded from further analysis\cite{zhao2023affective,lambert2024comparison}. 
\section{Motivation: Visualization of Stress Sensing Heterogeneity}
\label{motivation}
\textcolor{black}{This section presents our preliminary analysis investigating the heterogeneity of stress-sensing features across different datasets utilized in this study.}

\textcolor{black}{For each of the $340$ statistical features extracted from the raw sensor data (discussed in Section \ref{sec:feature}), we build an LMM model to assess group-wise differences while accounting for repeated measures within individuals. Given that stress responses vary both within and across individuals~\cite{nagaraj2023dissecting}, we included the \textit{user} identifier as a random effect to model inter- and intra-subject variability. Accounting for user-level random effects helps control for baseline variability between participants, improves the accuracy of fixed-effect estimates, and reduces confounding from subject-specific variation.}

\textcolor{black}{Group membership (e.g., Control, Pre-dose, Post-dose) was treated as a categorical fixed effect. Since physiological responses may differ by task condition (e.g., calm vs. stress-inducing tasks)~\cite{schmid2024individual}, we fit separate models for each task condition. This stratification allowed us to \textit{isolate task-specific effects} and better understand whether distributional shifts in features were consistent across contexts.}


\textcolor{black}{To systematically evaluate the group-wise differences, we conducted pairwise group comparisons using the following group contrasts:
\begin{enumerate}
    \item Control (reference) vs Pre-dose (comparison)
    \item Control (reference) vs Post-dose (comparison)
    \item Pre-dose (reference) vs Post-dose (comparison)
\end{enumerate}
}
\textcolor{black}{For each feature and comparison, we formulated the statistical hypothesis as follows:
\begin{itemize}
    \item H0: There is no significant difference in the feature distribution between the two groups, after adjusting for user-level random effects.
    \item H1: There is a statistically significant difference in the feature distribution between the groups, after adjusting for user-level random effects.
\end{itemize}
}
\textcolor{black}{The LMMs are formalized with Equation \ref{lmm_eqn}:
}
\begin{equation}
\label{lmm_eqn}
  \textcolor{black}{  f \sim Group + (1/user) }
\end{equation}

\textcolor{black}{Here, $f$ represents one of the 340 features. We perform a pairwise group comparison between a reference group $G_{0}$ and a comparison group $G_{1}$, where $Group$ is a categorical variable such that: $Group \in \{G_{0}, G_{1}\}$. The $user$ identifier is included as a random effect to account for subject-level variability.}


\textcolor{black}{We report the p-values (rounded to 3 decimal places) for each group comparison, corresponding to features which showed statistical significance, i.e, $p<0.05$. Detailed results, including coefficient estimates, standard errors, and 95\% confidence intervals, are provided in Appendix \ref{LMM_results_appendix}. The results corresponding to all the 340 features are available in our anonymous github repository provided in Appendix \ref{appendix:code}  . The following sections highlight key insights from each pairwise comparison.}

\begin{table}[h!]
\centering
\begin{subtable}[t]{0.35\textwidth}
\centering
\resizebox{\textwidth}{!}{\color{black}
\begin{tabular}{|c|c|c|}
\hline
\textbf{\makecell{Feature \\ name}} & \textbf{\makecell{Calm Task \\ p-value}} & \textbf{\makecell{Stress Task \\ p-value}} \\
\hline
L2-norm HR Kurtosis & 0.002 & 0.000 \\
HR Kurtosis & 0.013 & 0.000 \\
Number of Peaks in HR  & 0.041 & 0.014 \\
Skin Temp Kurtosis & 0.004 & 0.000 \\
HR Skewness & 0.032 & 0.005 \\
Number of Peaks in L2-norm Skin Temp  & 0.002 & 0.026 \\
95th Percentile L2-norm Tonic EDA & 0.001 & 0.007 \\
L2-norm Phasic EDA root mean square & 0.010 & 0.021 \\
Tonic EDA Shannon Entropy & 0.018 & 0.018 \\
Sum of L2-norm Skin Temp Differences & 0.033 & 0.029 \\
Skin Temp Min Value & 0.027 & 0.001 \\
Phasic EDA Kurtosis & 0.010 & 0.010 \\
Skin Temp Skewness & 0.001 & 0.036 \\
HR Count Below Mean & 0.000 & 0.002 \\
Skin Temp Count Below Mean & 0.016 & 0.000 \\
L2-norm HR Skewness & 0.029 & 0.003 \\
L2-norm Tonic EDA Kurtosis & 0.017 & 0.009 \\
\hline
\end{tabular}
}
\caption{\textcolor{black}{Control vs. Pre-dose}}
\label{tab:common_features_pre_dose_control}
\end{subtable}
\begin{subtable}[t]{0.35\textwidth}
\centering
\resizebox{\textwidth}{!}{\color{black}
\begin{tabular}{|c|c|c|}
\hline
\textbf{\makecell{Feature \\ name}} & \textbf{\makecell{Calm Task \\ p-value}} & \textbf{\makecell{Stress Task \\ p-value}} \\
\hline
HR Count Below Mean & 0.000 & 0.001 \\
Sum of L2-norm BVP & 0.002 & 0.045 \\
Sum of Squared BVP & 0.002 & 0.045 \\
Skin Temp Min & 0.007 & 0.001 \\
Skin Temp Count Below Mean & 0.009 & 0.001 \\
\hline
\end{tabular}
}
\caption{\textcolor{black}{Control vs. Post-dose}}
\label{tab:common_features_control_postdose}
\end{subtable}
\caption{\textcolor{black}{Common significant features across both calm and stress tasks for self-collected datasets.}}
\label{tab:common_features_combined}
\end{table}

\subsection{Insights from Statistical Analysis of Self-Collected Datasets}
\label{insights}

\textcolor{black}{This section first gives a brief overview of the above-mentioned pairwise group comparison analysis, and later discusses the corresponding identified insights.}

\begin{enumerate}
    \item \textcolor{black}{\textbf{Control vs Pre-dose} In the control vs. pre-dose comparison among the features analyzed, 46 features exhibited statistically significant group differences under the calm task, while 40 features were significant under the stress task. Importantly, 17 features were found to be significant across both task conditions, identifying them as \textit{potential} task-invariant physiological markers that distinguish the pre-dose group from the control group. The 17 common features across the different tasks are summarized in Table \ref{tab:common_features_pre_dose_control}, spanning multiple physiological domains, including heart rate (HR), temperature (TEMP), skin temperature (Skin Temp), heart rate variability(HRV), accelerometer (ACC), blood volume pulse(BVP), and electrodermal activity (EDA). The BVP signal is the primary output of Empatica E4 PPG sensor.}

    \item\textcolor{black}{ \textbf{Control vs Post-dose} For control vs. post-dose, 14 features showed statistically significant group differences under the calm task, while 11 were significant under the stress task. Of these, 5 features demonstrated consistent significance across both task contexts, indicating their potential task-invariant effects of the post-dose condition. They are listed in Table~\ref{tab:common_features_control_postdose}.}

    \item \textcolor{black}{\textbf{Pre-dose vs Post-dose} For the pre-dose vs. post-dose comparison, a total of 5 features showed significant differences between groups in the calm condition, while 12 features reached significance in the stress condition. However, only one feature, Phasic peaks, exhibited significant group differences across both tasks, suggesting a difference in modulation of phasic activity post-dose irrespective of task conditions.}
    
\end{enumerate}

\textcolor{black}{Our statistical analysis yielded the key observation that motivated the design choices of our work:}
\par 
\textcolor{black}{There is notable variability in the number of significant features across group comparisons: 69 for Control vs Pre-dose, 20 for Control vs Post-dose, and 16 for Pre-dose vs Post-dose, overall across the calm and stress tasks combined. Each comparison showed a different set of significant features. For example, BVP features like BVP sum and BVP energy were task-invariantly significant in Control vs Post-dose but were not for Control vs Pre-dose comparison. Even within the same group comparison, task-wise variability was observed. For example, between Pre-dose vs. Post-dose comparison, only one feature, \textit{Phasic peaks}, was significant across both calm and stress tasks, and the 15 features were significant only for one task. These findings suggest the existence of heterogeneous group-wise distribution shifts across tasks or contexts.}

\textcolor{black}{Notably, to explore domain generalization through feature selection, we identified all features that were consistently non-significant across comparisons (n = $257$), treating them as potentially distributionally robust. However, when we used these features to train a stress detection model (discussed in Section \ref{feature-select-discussion}), its performance was inferior to both OOD-specific baselines and our proposed HHISS method. This indicates that simple robust feature selection is insufficient, and there is a need to learn distributionally robust embeddings directly to better handle such distributional variability in stress detection.}

\par
\textcolor{black}{This finding motivated the development of our domain generalization framework, HHISS, which focuses on learning context-robust representations (i.e., robust across tasks and distributions) that avoid overfitting to condition-specific artifacts.}


\subsection{Comparison with Other Public Datasets}
\begin{table}[h!]
\centering
\begin{subtable}[h!]{0.35\textwidth}
\centering
\resizebox{\textwidth}{!}{\color{black}
\begin{tabular}{|c|c|c|}
\hline
\textbf{\makecell{Feature \\ name}} & \textbf{\makecell{Calm Task \\ p-value}} & \textbf{\makecell{Stress Task \\ p-value}} \\
\hline
ACC y Interquartile Range & 0.002 & 0.029 \\
ACC y Kurtosis & 0.026 & 0.035 \\
ACC y SVD Entropy & 0.001 & 0.031 \\
ACC y Perm Entropy & 0.001 & 0.031 \\
ACC z SVD Entropy & 0.001 & 0.031 \\
ACC z Max & 0.002 & 0.029 \\
ACC z Perm Entropy & 0.001 & 0.031 \\
ACC z Interquartile Range & 0.002 & 0.029 \\
ACC x Kurtosis & 0.026 & 0.035 \\
ACC L2 Max & 0.002 & 0.029 \\
Number of Peaks in HR  & 0.0013 & $\approx$0.000 \\
HR Kurtosis & 0.012 & $\approx$0.000 \\
HR Entropy & 0.005 & 0.004 \\
HR L2 Entropy & 0.035 & 0.032 \\
Root Mean Square of HR  & 0.014 & 0.011 \\
HRV Total Power & $\approx$ 0.000 & 0.008 \\
HRV Mean NN Interval & $\approx$ 0.000 & 0.008 \\
HR 95th Percentile & 0.015 & 0.013 \\
Root Mean Square of Skin TEMP  & 0.004 & 0.003 \\
\hline
\end{tabular}
}
\caption{\textcolor{black}{Control vs WESAD}}
\label{tab:common_features_control_wesad}
\end{subtable}
\begin{subtable}[h!]{0.3\textwidth}
\centering
\resizebox{\textwidth}{!}{\color{black}
\begin{tabular}{|c|c|c|}
\hline
\textbf{\makecell{Feature \\ name}} & \textbf{\makecell{Calm Task \\ p-value}} & \textbf{\makecell{Stress Task \\ p-value}} \\
\hline
ACC L2 Line Integral & $\approx$ 0.000 & 0.003 \\
ACC z 95th Percentile & $\approx$ 0.000 & 0.030 \\
ACC y 5th Percentile & $\approx$ 0.000 & 0.030 \\
ACC z Line Integral & $\approx$ 0.000 & 0.003 \\
BVP Energy & 0.001 & $\approx$ 0.000 \\
BVP Skewness & 0.024 & 0.004 \\
BVP Sum & 0.050 & $\approx$ 0.000 \\
L2-norm HR Skewness & $\approx$ 0.000 & $\approx$ 0.000 \\
TEMP Count Below Mean & 0.040 & $\approx$ 0.000 \\
\hline
\end{tabular}}
\caption{\textcolor{black}{Control vs AffectiveRoad}}
\label{tab:common_features_control_affectiveroad}
\end{subtable}
\begin{subtable}[H]{0.27\textwidth}
\centering
\resizebox{\textwidth}{!}{\color{black}
\begin{tabular}{|c|c|c|}
\hline
\textbf{\makecell{Feature \\ name}} & \textbf{\makecell{Calm Task \\ p-value}} & \textbf{\makecell{Stress Task \\ p-value}} \\
\hline
HR Kurtosis & 0.037 & 0.001 \\
BVP L2 Kurtosis & 0.029 & 0.020 \\
L2-norm HR Peak-to-Peak & 0.043 & 0.003 \\
 Skin Temp Root Mean Square & 0.048 & 0.011 \\
BVP L2 Interquartile Range & 0.003 & 0.041 \\
Number of Peaks in Tonic EDA  & 0.012 & 0.009 \\
HRV pNN20 & 0.041 & 0.014 \\
\hline
\end{tabular}}
\caption{\textcolor{black}{WESAD vs AffectiveRoad}}
\label{tab:common_features_wesad_affectiveroad}
\end{subtable}
\caption{\textcolor{black}{Common significant features across both calm and stress tasks for publicly available datasets.}}
\label{tab:common_features_public_control}
\end{table}

\label{lmm_public_data}
\textcolor{black}{To better understand the generalizability of our findings and the physiological feature shifts across other data distributions, we extended our statistical analyses to two other publicly available datasets: WESAD and AffectiveRoad, described in Section~\ref{dataset:public}. Since these datasets comprise only healthy participants, comparisons were made against our control group data. The feature extraction process for the WESAD and AffectiveRoad remained the same as discussed in Section \ref{sec:feature}. Our analyses yielded the following findings:}

\begin{enumerate}
    \item \textcolor{black}{\textbf{Control vs WESAD:} Out of the 340 features analyzed, 55 features exhibited statistically significant differences under the calm condition and 33 under the stress condition. Nineteen features were consistently significant across both task contexts (Table \ref{tab:common_features_control_wesad}).}
    \item \textcolor{black}{\textbf{Control vs AffectiveRoad:} This comparison revealed fewer consistent patterns. Among the features evaluated, 30 were statistically significant under calm and 46 under stress, with only 9 features significant across both conditions (Table \ref{tab:common_features_control_affectiveroad}).}
    \item \textcolor{black}{\textbf{WESAD vs AffectiveRoad:} Between WESAD and AffectiveRoad, 30 features showed significance in the calm task and 58 under stress. Seven features were found to be consistently significant in both tasks (Table \ref{tab:common_features_wesad_affectiveroad}).}
\end{enumerate}




\textcolor{black}{Overall, the patterns observed across these publicly available datasets align with the insights identified for our self-collected datasets (discussed in Section \ref{insights}). These observations further validate the existence of heterogeneous group-wise distribution shifts across tasks or contexts.}

\section{HHISS Framework}\label{our-approach}
This section discusses the HHISS framework and its design choices, supported by Algorithm \ref{alg:1} and Figure \ref{fig:workflow}.

\emph{Overview:} The HHISS framework addresses the heterogeneity in human physiological stress responses through four key components outlined in Figure \ref{fig:workflow}. 
\textcolor{black}{\emph{First}, it incorporates IRM-based training regularization by treating each individual as a separate domain. This guides the training of an IRM-regularized over-parameterized embedding generator network that aims to learn causally invariant features that generalize across individuals for stress detection through the follow-up decision-making network. However, such IRM-regularized over-parameterized  networks are prone to overfitting, which can hinder generalization.}
This is typically mitigated by employing sparse training alongside IRM \textcolor{black}{regularization}, but sparse training can introduce bias by favoring dominant individuals during pruning. The \emph{second} component enhances generalizability and minimizes bias by presenting a novel subject-wise gradient-based pruning and identifying the intersection of pruned sub-networks. This ensures that the retained parameters are consistent across individuals, capturing invariant features critical for generalization. Despite these efforts, the pruning intersection process may result in a significantly reduced sub-network, limiting the model's capacity to retain essential knowledge. To address this, the \emph{third} component introduces sparse-to-sparse training, which iteratively refines the intersectional sub-network by allowing parameters to relearn and progressively evolve toward a larger, more effective substructure. Finally, the \emph{fourth} component introduces a novel approach to enhancing generalizability further, a critical step since sparse-to-sparse iterative training can overfit on limited human sensing data. This component leverages the continuous labels from the \textcolor{black}{IRM-regularized over-parameterized model (complete model, including the embedding and decision-making network layers)} obtained through the first component, encouraging a flatter loss landscape and ensuring robust performance across diverse populations and unseen conditions.

Each of the four competent is detailed and justified below:




\begin{figure}[h!]
    \centering
    \includegraphics[width=0.7\linewidth]{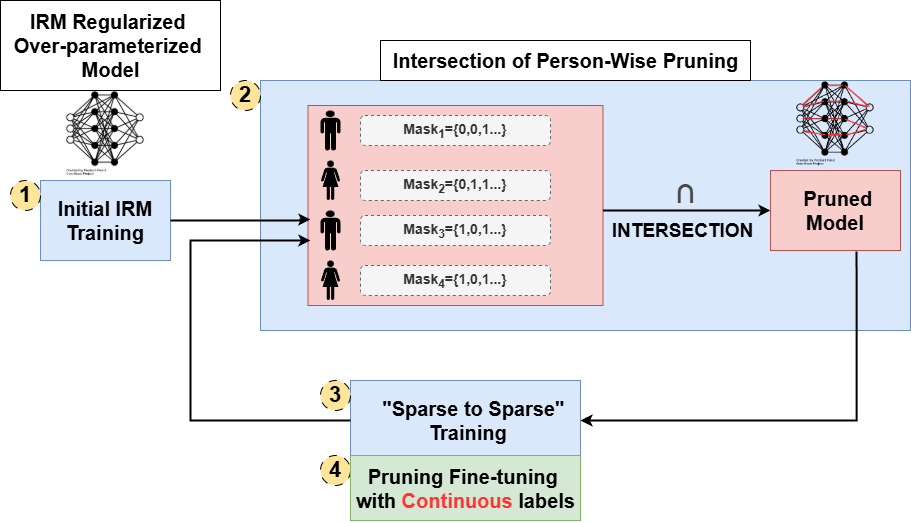}
    \caption{Approach Workflow}
    \label{fig:workflow}
    \vskip -3ex
\end{figure}

\subsection{\textcolor{black}{Causally Invariant Features Extraction through Initial IRM Regularization:} }\label{Algo:IRM-training-1}
\subsubsection{Challenge:}
Human physiological responses to stressors are heterogeneous across inter- \cite{lecic2011stress,gillespie2009risk,jacoby2021individual} and intra- individuals \cite{nagaraj2023dissecting}. To develop a robust model applicable across diverse scenarios, it is crucial to ensure the model focuses on invariant and consistent features or embedding generation.

\subsubsection{Our Solution:} 
To address the challenge, the HHISS framework uses Invariant Risk Minimization (IRM) \textcolor{black}{regularization as the first training step. IRM \cite{arjovsky2019invariant} is a regularization strategy during training that encourages models to learn invariant features with a stable, causal relationship with the target variable (i.e., stress assessment) across all training domains; hence, such features are more generalizable to unseen domains. Notably, it is compatible with any downstream decision-making network architecture.} In this work, focused on human-centered stress sensing, we consider each subject to be a unique domain.
As outlined in line $1$ of Algorithm \ref{alg:1} (and Figure \ref{fig:workflow}(1)), following the methodology of \citet{arjovsky2019invariant}, we train a neural network using the IRM loss function $L_{IRM}$ (equation 6 in \cite{arjovsky2019invariant}). The resulting \textcolor{black}{IRM-regularized over-parameterized embedding generation model, $f_{\theta}$, is designed to learn a diverse causally invariant feature representation across all individuals in the training set, enhancing its ability to generalize to new individuals and stressors.}

\begin{algorithm}
\caption{Approach}
\label{alg:1}
\begin{algorithmic}[1]

\REQUIRE Given: \\$S_{1},S_{2}...S{n} \in S $:  \text{Data belonging to n known subjects}\\
$R$: \text{Number of training rounds}\\
$T$: \text{Sparse-to-sparse threshold}\\
$K$: \text{Pruning amount}\\

\STATE $f_{over}(\cdot|\theta )$: $\text{Over-parameterized model trained with IRM}$  
\STATE $f_{pruned}(\cdot| \theta ) \leftarrow f_{over}(\cdot|\theta )$ \\
\REPEAT
\FOR{$S_{i}$ in $S$}
\STATE$\text{Mask}_i = Prune( f_{pruned}(\cdot| \theta ), S_i,K) $
\ENDFOR
\STATE$ Mask = \cap_i Mask_i$\\
\STATE$ \text{Apply } Mask \text{ to the model } f_{pruned}(\cdot| \theta )$\\

\REPEAT
\STATE$Loss =\sum_{S_i \in S} L_{IRM} + \lambda L_{Continuous}$
\STATE$\text{Update the model with Loss}$
\UNTIL{Model performance higher than T}
\UNTIL{Repeated for R times}
\vskip -5ex
\end{algorithmic}
\end{algorithm}

\subsection{Intersection of Person-wise Sub-networks to Enhance Generalizability: }\label{approach-intersection-pruning}
Previous research has shown that the generalization ability of \textcolor{black}{IRM-regularized embeddings} can be compromised by overfitting, particularly in \textcolor{black}{IRM-regularized over-parameterized networks} \cite{zhou2022sparse}. 
IRM regularization term promotes generalizability. However, overfitting reduces gradient values, resulting in a reduction of the magnitude of the IRM regularization term, thus adversely impacting the generalizability. 

The HHISS framework relies on \textcolor{black}{IRM-regularized over-parameterized} networks in its initial step since it provides the capacity to capture a wide range of potential patterns and relationships in the data \cite{zhang2021understanding}. This is crucial for human-centered stress sensing, where physiological responses exhibit significant heterogeneity.

To address the challenges posed by overparameterization, prior studies have employed sparse training techniques alongside \textcolor{black}{IRM regularization} to enforce the learning of invariant features and prevent overfitting \cite{zhou2022sparse}. Moreover, introducing sparsity can further enhance the model's generalization by improving training and regularization, mitigating the impact of noisy examples, and resulting in a flatter loss landscape \cite{bartoldson2020generalization,jin2022neural}.

\subsubsection{Challenge:} 
\begin{figure}[h!]
    \centering
    \includegraphics[scale=0.38]{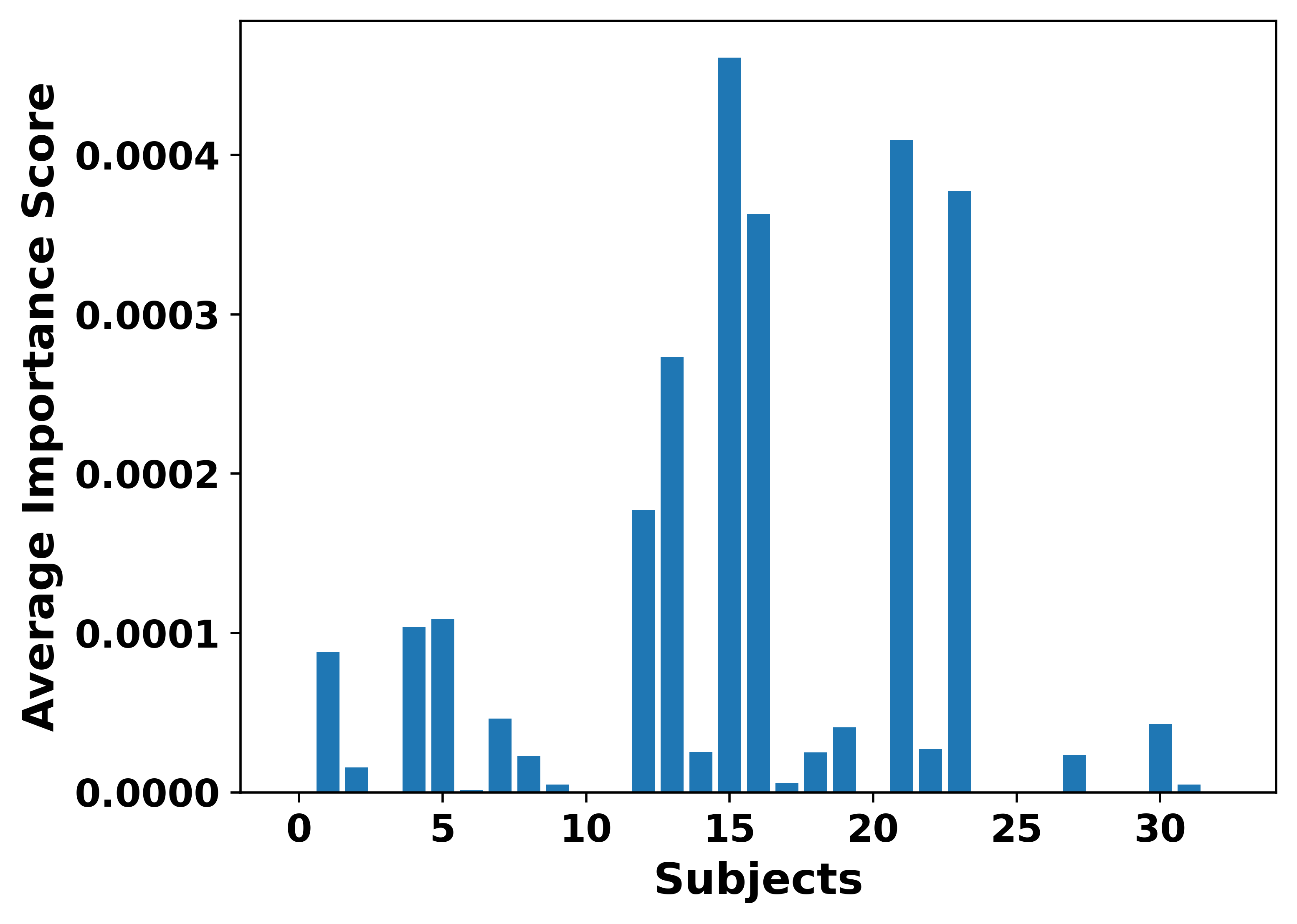}
    \caption{Subject wise importance score on the non-subject wise gradient based pruning.}
    \label{fig:imp}
    \vskip -3ex
\end{figure}

However, the identification of sparse networks using collective stress data can be influenced by certain groups of individuals, thus resulting in a loss of generalizability. To explore this issue, we conduct gradient-based pruning \cite{singh2020woodfisher} on an IRM-regularization trained neural network to identify the resulting pruned sub-network. In Figure \ref{fig:imp}, we visualize the average importance score (a metric used to assess the significance of neurons for model prediction) of the preserved neurons in the pruned sub-network for each individual subject. The results indicate that certain subjects exhibit significantly higher importance scores due to their elevated gradients. As a result, neuron selection and, consequently, sub-network selection during pruning are heavily influenced by these subjects. This suggests that the characteristics or features most prominent in these individuals are retained in the pruned sub-network, potentially leading to the exclusion of other subjects' characteristics and resulting in biased model predictions for unseen individuals.

\subsubsection{Our Solution:}
\label{section:person-wise-pruning-solution}
As outlined in the lines $4-8$ of Algorithm \ref{alg:1} (and Figure \ref{fig:workflow}(2)), we address this human-centered challenge by performing subject-wise pruning for each individual and identifying the intersection as the subject-invariant pruned sub-network. Specifically, we apply gradient-based pruning to determine the set of important parameters for each subject, resulting in $n$ model substructures. To consolidate the subject-invariant information contained within these subject-specific substructures, we compute their intersection. 
Parameters lying outside the intersection set carry varying importance across subjects and hence may introduce bias and were subsequently zeroed out. Consequently, \emph{the remaining non-zero parameters in the model represent those with consistently high importance across all subjects, ensuring subject-invariant significance.}

The intersection \emph{`Mask'} can be formulated as:


\begin{equation}
\begin{aligned}
Mask &= \bigcap_{i}^{S_i \in S} Mask_i, Mask_i = \{0,1\}^J
\end{aligned}
\end{equation}

Here, the \emph{`Mask'} is the final mask used for model pruning, which is the intersection of all subjects' masks. $Mask_{i}$ represent the binary pruning mask for subject $i$. It is a binary mask array length of J, where each element can be either 0 or 1. Each weight in the neural network has a corresponding mask value $m_{j}$, where $m_{j}=1$ for retained weights and $m_{j}=0$ for pruned weights. 
The mask value $m_{j}$ is determined by an importance score. We retain the top(100-K)\% of weights and prune the lowest K\%, where pruning amount K is a tunable hyperparameter. The importance score follows the gradient-based pruning approach \cite{kang2023fashapley} and is computed using Equation \ref{eq:importance} as the absolute value of the gradient of the loss with respect to the weights, multiplied by the absolute value of the weight.

\begin{equation}
\begin{aligned}
\text{importance score}= | \frac{\nabla L}{\nabla \theta}  |  | \theta| \quad
\end{aligned}
\label{eq:importance}
\end{equation}




\subsection{“Sparse-to-sparse” Training as Pruning Fine-tuning to Attain an Effective Intersectional Sub-network}\label{our-approach-DST}

Conventionally, pruning is followed by a retraining step over the non-pruned parameters, i.e., the non-zero sub-network (known as fine-tuning) that helps to relearn the information lost by removal of the pruned or zeroed-out parameters \citep{ThiNet-ICCV17, zhu2017prune,GP-soa}.

\subsubsection{Challenge:} Given the heterogeneity of human psychophysiological responses to stress, performing the intersectional `Mask' identification as described in Section \ref{section:person-wise-pruning-solution} as a one-time process results in a significantly reduced number of non-zero parameters (or sub-network) compared to the regular pruning with same pruning amount $K$. This reduction eliminates a substantial portion of the learned knowledge crucial for effective model generation. Furthermore, this minimal substructure contains insufficient parameters to relearn a generalizable and effective model during fine-tuning. A more detailed discussion of this challenge, along with parameter visualization, is provided in Section \ref{Sparse-to-sparse}.

\subsubsection{Our Solution:} To address this, it is crucial to learn an intersectional sub-network that is larger, preserving enough learned knowledge from the over-parameterized neural network and retaining sufficient parameters to recover the lost information through zeroed-out parameters to attain a generalizable and effective model during fine-tuning.
\par 
We address this issue through `sparse-to-sparse' training \cite{liu2023don, dai2019nest}, as outlined in the loop in lines $3-13$ of Algorithm \ref{alg:1} (and Figure \ref{fig:workflow}(3)). After the initial intersectional `Mask' identification and zeroing out other parameters (line 8 of Algorithm \ref{alg:1}), the model undergoes further training, allowing all parameters, including the zeroed-out ones, to learn until the model attains a threshold efficacy of $T$. This process is then repeated, involving re-identification of the intersectional `Mask' and re-training of all parameters on the relearned model for subsequent rounds.
\par 
The rationale for using `sparse-to-sparse' training, supported by literature \cite{dai2019nest}, is that, in the initial rounds, it identifies evolving or varying sub-networks but eventually converges to a consistent one. Consequently, in HHISS, as the rounds progress, the subject-wise pruned network for each individual converges towards similarity, resulting in a consistently larger intersectional `Mask' that is sufficiently sized and retains enough information to develop a generalizable and effective model.

\subsection{Avoid Overfitting by Leveraging Continuous Labels}\label{our-approach-KD}

\par 
\subsubsection{Challenge:} The consistent intersectional `Mask' identification requires multiple rounds of `sparse-to-sparse' training. Furthermore, human-centered datasets, particularly in the health domain (e.g., data from the substance abuse population), are limited in size \cite{maxhuni2017managing,yfantidou2023beyond}. Such factors may lead to the model being overfitting and lose generalizability. 

\subsubsection{Our Solution:} To prevent overfitting, which results in the model converging to a narrow point in the loss landscape, we leverage recent literature \cite{zhang2023generalization, wang2021embracing} that suggests models trained with continuous labels tend to converge to a flatter loss landscape, corresponding to improved generalizability.

In the HHISS approach, continuous labels are derived directly from \textcolor{black}{the IRM-regularized over-parameterized model (end-to-end model including the embedding and decision-making layers)} obtained in Section \ref{Algo:IRM-training-1}. Unlike knowledge distillation (KD)-based methods that typically use temperature scaling to smooth predicted logits and transfer soft information between teacher and student models \cite{zhang2023generalization}, HHISS entirely avoids temperature scaling. Instead, it focuses on preserving the original predictive structure of the logits. The objective is to utilize continuous labels, rather than discrete ones (stress vs. calm), without mimicking the behavior of the parent network (in this case, the IRM-regularized over-parameterized model), which is typical in KD-based techniques \cite{zhang2023generalization,chai2022fairness,wang2021embracing}. Mimicking the parent network could compromise the ability to extract subject-invariant features, a critical capability achieved through steps 2 in Figure \ref{fig:workflow}, as discussed earlier. This distinction underscores the unique way HHISS handles continuous logits by retaining their original information.

\par 
Hence, as outlined in lines $10-11$ of Algorithm \ref{alg:1} (and Figure \ref{fig:workflow}(4)), we fine-tune utilizing a $Loss$ that includes utilizing a continuous label we attain through the logit values obtained from the IRM-regularized over-parameterized model \textcolor{black}{(end-to-end network)} $f_{over}(\cdot|\theta )$.

The fine-tuning loss function can be summarized as follows:

\begin{equation}
\begin{aligned}
    \text{Loss} &= \sum_{S_i \in S} L_{IRM} + \lambda L_{Continuous}, \\
    L_{Continuous} &= L_{CN}\big(f_{pruned}(X^{i}), f_{over}(X^{i})\big)
\end{aligned}
\label{eq:finetune}
\end{equation}

Here, $L_{CN}$ is cross entropy loss. The $L_{IRM}$ is the IRM loss defined by \cite{arjovsky2019invariant}. 
\par 
Consequently, through $L_{Continuous}$, HHISS further promotes generalization by exploiting continuous labels ($f_{over}(X^{i})$: logit from $f_{over}(\cdot|\theta )$) and rich knowledge learned by the over-parameterized model, resulting in a flatter loss landscape \cite{zhang2023generalization,wang2021embracing}. This is further discussed in Section \ref{loss-landscape-perspective}, with reference to Figure \ref{fig:loss_landscape}.

\section{Experiment setup}\label{experiment-setup}

The following sections provide a comprehensive evaluation of our proposed approach, assessing the overall method's performance and its individual components. We benchmark our results against state-of-the-art DG and DA approaches. The presented network parameter configurations were optimized by performing a grid search of the possible network parameters. The evaluation results presented are end-to-end, incorporating the inaccuracy due to pre-processing errors. \textcolor{black}{We use balanced accuracy and macro F1 score as evaluation metrics. Balanced accuracy calculates the average recall (true positive rate) across all classes, ensuring that each class contributes equally regardless of its frequency. Similarly, the macro F1 score computes the F1 score for each class independently and then averages them. Both metrics are well-suited for imbalanced datasets , as they treat all classes equally\cite{zhang2024reproducible}.}
For each dataset and approach, we use neural network architectures previously benchmarked and deployed for similar datasets, without considering variations in architecture \cite{li2020stress,dziezyc2020can}.

\subsection{Baseline OOD approaches}
\textcolor{black}{As a simple random baseline benchmark, we use a majority vote classifier, which always predicts the most frequent class in the dataset.\cite{zhang2024reproducible}}. As a non-OOD approach baseline, we considered Empirical Risk Minimization (ERM), a fundamental principle in machine learning that aims to find the optimal model parameters by minimizing the empirical risk on the training data. Here, we use cross-entropy loss to optimize the model.

 To demonstrate the OOD generalization or DG superiority of HHISS, as mentioned in Section \ref{rel-work-DG}, our baselines include Invariant Risk Minimization (IRM) \cite{arjovsky2019invariant}, Distributionally Robust Optimization (DRO)\cite{sagawa2019distributionally},  Risk Extrapolation (Vrex)\cite{krueger2021out}, \textcolor{black}{Clustering\cite{xu2023globem}, Domain adversial Neural network (DANN) \cite{ajakan2014domain,meegahapola2024m3bat}, Meta-Learning for Domain Generalization(MLDG) \cite{li2018learning}}, and sparse training with IRM \cite{zhou2022sparse}, referred to as SparseTrain.
Furthermore, we evaluated gradient magnitude pruning \cite{granger2021progressive}, referred to as ERM pruning, and knowledge distillation (KD)\cite{reddy2024towards}. 
\par 
Moreover, as a recent unsupervised learning DA baseline, we include Test-Time Personalization (TTP) \cite{xu2023joint}.

\subsection{Dataset Setup}
\label{dataset_setup}
 To evaluate the robustness of the binary stress detection model trained on our collected control dataset, we employed a person-disjoint hold-out method \cite{zhang2024reproducible}. From the self-collected healthy control group (Section \ref{self-dataset-discussion}), 32 subjects were randomly selected for training, referred to as the Control-training set, while 14 subjects were reserved for evaluation, referred to as the Control-testing set. In the remaining datasets, all subjects were used exclusively for evaluation, ensuring a thorough assessment of the model's OOD robustness across diverse subjects and conditions. \textcolor{black}{This setup was maintained in all evaluations except for Sections \ref{RQ-2-other-datasets-training}, \ref{rq5}, and \ref{rq4}, which are discussed below.}

\textcolor{black}{We applied the similar person-disjoint hold-out strategy in Sections \ref{rq4} and \ref{rq5}. However, in Section \ref{rq4}, 12 randomly selected WESAD subjects were used for training, while all remaining data (including other datasets) were used for testing. In Section \ref{rq5}, randomly selected 12 subjects from WESAD and 32 from the control group were used for training, with all remaining data reserved for evaluation.}

\begin{figure}
    \centering
    
    \includegraphics[width =0.5\linewidth]{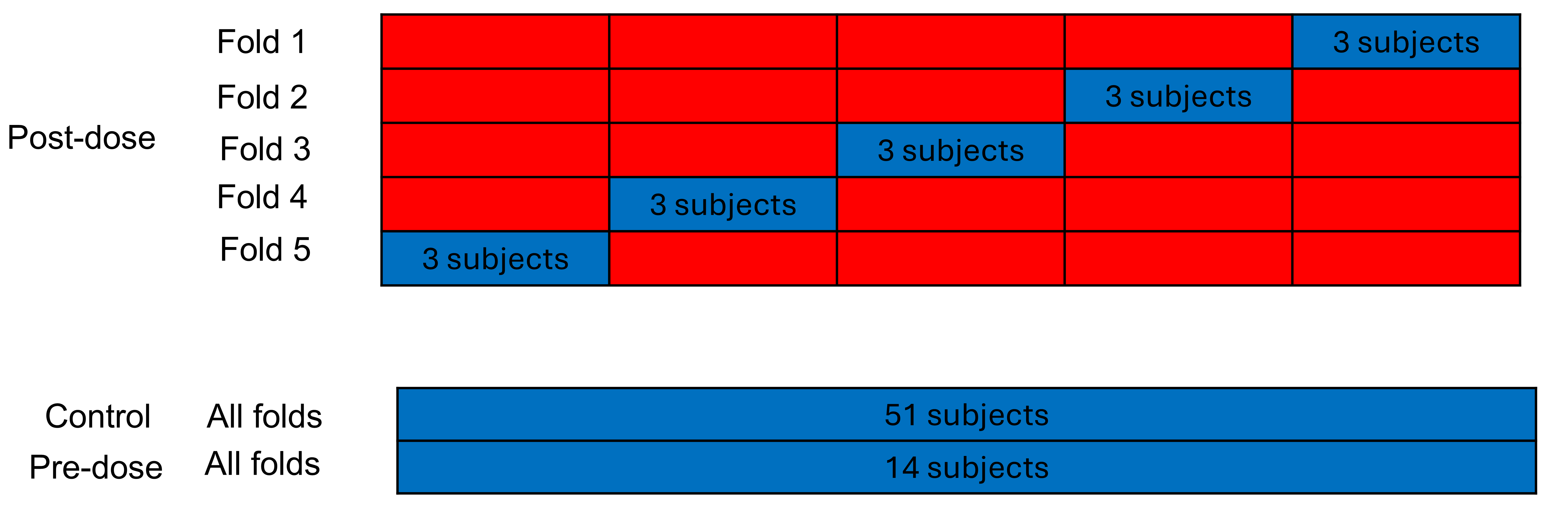}

    \caption{5-fold Subject Independent Cross-Validation}
\vskip -3ex
    \label{fig:datasplit}
 
\end{figure}

\textcolor{black}{In contrast to other evaluations, Section \ref{RQ-2-other-datasets-training} evaluates models trained on the Pre-dose and Post-dose datasets, which contain only 14 and 15 sessions, respectively. To mitigate this data limitation, a 5-fold cross-validation strategy was employed, ensuring person-disjoint train-test splits. Utilizing Figure \ref{fig:datasplit}, the strategy is demonstrated while training was performed on the Post-dose dataset. It was divided into five groups; in each fold, one group was used for testing and the rest for training. Additionally, all participants from the Control and Pre-dose datasets were held out for testing in every fold. In Figure \ref{fig:datasplit}, subjects in red represent training data, while those in black represent evaluation data. This setup ensured a relatively larger training set in each fold.}

\section{Evaluation of HHISS}

This section highlights the generalizability and persistence of the HHISS approach by comparing it with baseline methods across four key questions. Following this, a comprehensive ablation study is performed to provide deeper insights into the performance of the HHISS approach. Finally, we offer empirical justifications for the design choices, shedding light on the factors contributing to its enhanced performance.

\subsection{Evaluating HHISS Generalizability and Adaptability in OUD and Beyond}\label{results-baselines}

In this section, we compare HHISS approach with baseline approaches discuss in Section \ref{experiment-setup} on the topic of four questions.

\subsubsection{Q1: How well can the model perform on the OUD population when trained exclusively on healthy control groups?}\label{rq1}

 As highlighted in the literature \citet{mathur2019unsupervised}, an effective sensing approach must be both persistent and robust, achieving high performance on both in-distribution and out-of-distribution scenarios.
 Table \ref{table:baseline} presents the performance results when trained on the Control-training set and evaluated on the Control-testing set as the in-distribution scenario and on the Pre-dose and Post-dose datasets as out-of-distribution scenarios. Results for each dataset are reported individually, along with aggregated OOD mean accuracy. 

Among the aggregated OOD results, the top-performing baselines—SparseTrain and ERM Pruning—achieved mean accuracies of approximately 69.95\%. In comparison, HHISS outperformed all methods with a mean OOD accuracy of 75\%, demonstrating significantly higher robustness in OOD scenarios.

For in-distribution results, HHISS achieved about 1–7\% higher accuracy compared to the baselines, showcasing state-of-the-art persistence. This indicates that HHISS not only achieves superior OOD robustness but also maintains similar or better performance on in-distribution data. Among the baselines, SparseTrain showed the best overall accuracy across all distributions. However, despite its comparable in-distribution accuracy to HHISS, SparseTrain's OOD mean accuracy was approximately 6\% lower, highlighting HHISS's clear superiority over SparseTrain. \textcolor{black}{It can be observed that clustering methods outperform the ERM baseline on F1 score in the Control group but fail in OOD scenarios. This highlights the heterogeneity between the OUD and Control subjects, making it difficult to assign OUD subjects to any Control group cluster.}

Notably, HHISS excels in OOD robustness under distribution shifts caused by various health conditions, such as Pre- and Post-dose OOD scenarios. However, as shown in Table \ref{table:baseline}, all methods face challenges with Pre-dose datasets. This poor performance is attributed to the higher deviation of stress patterns exhibited by Pre-dose participants (Section \ref{motivation}), even during calm conditions or tasks, which complicates OOD prediction when models are trained exclusively on healthy controls. 


In summary, HHISS not only maintains comparable performance on in-distribution datasets but also significantly boosts performance in out-of-distribution (OOD) scenarios, specifically the OUD population dataset. Its ability to consistently outperform other methods, particularly in OOD settings, demonstrates its OOD robustness across a variety of conditions. This makes HHISS a reliable model for real-world applications involving the OUD population.


 
\begin{table*}[]
\resizebox{0.85\linewidth}{!}{
\begin{tabular}{l|ll|lllll}
\hline
            & \multicolumn{2}{l|}{In distribution} & \multicolumn{5}{l}{Out of distribution}                                                                                    \\ \hline
            & \multicolumn{2}{l|}{Control}         & \multicolumn{2}{l}{Pre-dose} & \multicolumn{2}{l}{Post-dose} &                   \multicolumn{1}{|l}{OOD Mean}                                          \\ \hline
Approach    & Accuracy          & Macro F1         & Accuracy      & Macro F1     & Accuracy      &Macro F1      & \multicolumn{1}{|l}{Accuracy}\\ \hline
\textcolor{black}{Majority Vote}         & 0.5             & 0.4305           & 0.5         & 0.4302        & 0.5        & 0.4366        &       \multicolumn{1}{|l}{0.5}                                                \\ \hline
ERM         & 0.767             & 0.7475           & 0.654         & 0.658        & 0.6884        & 0.7083        &       \multicolumn{1}{|l}{0.6712}                                                \\ \hline

\textcolor{black}{Feature Select}        &   0.781         &    0.756       &    0.6278      &    0.6348    &   0.6847     &    0.6951   &       \multicolumn{1}{|l}{0.6562}                                                \\ \hline

IRM         & 0.7942            & 0.7781           & 0.6751        & 0.6723       & 0.7035        & 0.6795        &        \multicolumn{1}{|l}{0.6893}                                               \\ \hline
Vrex        & 0.7738            & 0.7456           & 0.6327        & 0.6435       & 0.7204        & 0.7267        &    \multicolumn{1}{|l}{0.6765}                                                  \\ \hline
DRO         & 0.7768            & 0.742            & 0.6597        & 0.669        & 0.7321        & 0.7341        &      \multicolumn{1}{|l}{0.6959}                                                 \\ \hline
TTP         & 0.7649            & 0.7328           & 0.6394        & 0.6418       & 0.7139        & 0.7043        & \multicolumn{1}{|l}{0.6766}                                                   \\ \hline
\textcolor{black}{DANN}         & 0.7703            & 0.7404           & 0.6717        & 0.6522       & 0.697        & 0.6591        & \multicolumn{1}{|l}{0.6843}                                                   \\ \hline
\textcolor{black}{MLDG}         &0.7953	&0.7755	&0.6543&	0.6658&	0.7179&	0.7314
        & \multicolumn{1}{|l}{0.6861}                                                   \\ \hline
\textcolor{black}{Clustering}          & 0.7609            & 0.7911           & 0.6589        & 0.6539       & 0.6799        & 0.6713        & \multicolumn{1}{|l}{0.6694}                                                   \\ \hline
ERM pruning & 0.7733            & 0.7173           & 0.674         & 0.6758       & 0.725         & 0.6854        &     \multicolumn{1}{|l}{0.6995}                                                  \\ \hline

SparseTrain & 0.8285            & 0.8236           & 0.6634        & 0.6827       & 0.7299        & 0.7538        &   \multicolumn{1}{|l}{0.6966}                                                   \\ \hline
KD          & 0.7863            & 0.764            & 0.6403        & 0.6429       & 0.7336        & 0.7231        &    \multicolumn{1}{|l}{0.6869}                                                  \\ \hline
HHISS       & 0.8371            & 0.8206           & 0.7181        & 0.717        & 0.7817        & 0.7942        & \multicolumn{1}{|l}{0.7500}                                                      \\ \hline
\end{tabular}
}
\caption{Baseline performance evaluated using balanced accuracy and macro F1 score}
\vskip -5ex
\label{table:baseline}

\end{table*}

\subsubsection{Q2: How does the heterogeneity in OUD data affect model training across different OUD conditions?}\label{RQ-2-other-datasets-training}

As established in Section \ref{motivation}, there is significant heterogeneity across OUD conditions (Pre-dose, Post-dose) and Control. This section evaluates how this heterogeneity influences the performance based on the dataset used for training. Two scenarios were considered: training on the Pre-dose dataset and training on the Post-dose dataset, with evaluations conducted on the Control-testing, Pre-dose, and Post-dose datasets. The results of training on the Pre-dose and Post-dose datasets are presented in Table \ref{tab:predose} and Table \ref{tab:postdose}, respectively.


When trained on the Pre-dose dataset, the HHISS model exhibited superior performance on the Pre-dose dataset itself, but comparing the results in Table \ref{table:baseline}, the performance on the Control and Post-dose datasets are lower.
However, among all approaches, HHISS achieved the best performance across all datasets. Notably, when evaluated on the Post-dose dataset, HHISS outperformed the best baseline with a 6\% improvement in accuracy. Similarly, as shown in Table \ref{tab:postdose}, when trained on the Post-dose dataset, the HHISS model achieved the highest accuracy on the Post-dose dataset but exhibited decreased accuracy on the Control-testing and Pre-dose datasets relative to Table \ref{table:baseline}. This trend was consistently observed across various baselines, including SparseTrain and ERM pruning approaches. Notably, HHISS outperformed all baselines, particularly on the Pre-dose dataset, achieving a 4\% improvement in accuracy compared to the best baseline.

These results show that training on the Pre-dose dataset reduces performance on the Post-dose dataset and vice versa, with a more significant decline compared to the Control-testing dataset. This reflects a greater heterogeneity between the Pre-dose and Post-dose datasets. For the Control-testing dataset evaluation, models trained on Post-dose data performed better than those trained on Pre-dose data. This suggests Post-dose data is more similar to Control data, aligning with observations in Section \ref{insights}.


This section's evaluations highlight that the two OUD conditions, Pre-dose and Post-dose, exhibit significant heterogeneity between each other while being relatively homogeneous within their own distributions. This homogeneity within Pre- and Post-dose datasets limits the diversity of stress patterns they incorporate, making them less effective for training OOD models. Consequently, models such as HHISS or baseline approaches trained on these datasets struggle to capture broader invariant stress patterns, leading to lower OOD robustness compared to models trained on the Control dataset. Our discussion in Section \ref{group-variability} further supports this observation.

This finding highlights an \emph{important insight:} 

Developing robust OOD stress detection models does not necessarily require data from a wide range of disease conditions, especially when collecting such data is infeasible. For instance, obtaining labeled data from all medication states in the OUD population is infeasible, and training with only Pre- or Post-dose data led to suboptimal results.
Instead, leveraging diverse Control data is an effective approach for achieving superior accuracy across all distributions. The proposed HHISS framework, with its human-centered novel design, enhances OOD stress detection robustness while utilizing only control data for training, making it especially valuable for health applications where collecting diverse datasets is challenging.

\begin{table}[h!]
    \centering
    \begin{minipage}{0.48\linewidth}
        \centering
        \resizebox{\linewidth}{!}{
        \begin{tabular}{l|l|ll|l}
\hline
            & In distribution & \multicolumn{2}{l|}{Out of distribution} & Overall  \\ \hline
Approach    & Pre-dose       & Control            & Post-dose            & Mean Acc \\ \hline
ERM         & 0.7454          & 0.7078             & 0.6255              & 0.6929   \\ \hline
IRM         & 0.7443          & 0.749              & 0.6662              & 0.7198   \\ \hline
Vrex        & 0.7791          & 0.7473             & 0.6158              & 0.714    \\ \hline
DRO         & 0.7467          & 0.767              & 0.6702              & 0.7279   \\ \hline
TTP         & 0.7407          & 0.7174             & 0.6407              & 0.6996   \\ \hline
\color{black} DANN         & 0.7157          & 0.6934              & 0.5934              &  0.6675  \\ \hline
\color{black} MLDG         & 0.7262          & 0.7283              & 0.6202              &  0.6917  \\ \hline

\color{black} Clustering & 0.7606          & 0.7035             & 0.6434              & 0.7025   \\ \hline
ERM pruning & 0.7911          & 0.7487             & 0.7037              & 0.7478   \\ \hline
KD          & 0.7322          & 0.7777             & 0.6372              & 0.7157   \\ \hline
SparseTrain & 0.773           & 0.7689             & 0.6657              & 0.7359   \\ \hline

HHISS       & 0.8013          & 0.7875             & 0.7683              & 0.7857   \\ \hline
\end{tabular}
        }
        \caption{Accuracy for Models Trained with Pre-Dose Data}
        \label{tab:predose}
    \end{minipage}%
    \hfill
    \begin{minipage}{0.48\linewidth}
        \centering
        \resizebox{\linewidth}{!}{
        \begin{tabular}{l|l|ll|l}
\hline
            & In distribution & \multicolumn{2}{l|}{Out of distribution} & Overall  \\ \hline
Approach    & Post-dose       & Control            & Pre-dose            & Mean Acc \\ \hline
ERM         & 0.7164          & 0.7395             & 0.6068              & 0.6876   \\ \hline
IRM         & 0.7517          & 0.7548             & 0.6322              & 0.7129   \\ \hline
Vrex        & 0.7096          & 0.7533             & 0.6356              & 0.6995   \\ \hline
DRO         & 0.7219          & 0.7496             & 0.6222              & 0.6979   \\ \hline
TTP         & 0.7383          & 0.723              & 0.6237              & 0.695    \\ \hline
\color{black} DANN         & 0.7413          & 0.7008             & 0.6424              & 0.6948   \\ \hline
\color{black} MLDG         & 0.7073          & 0.7114             & 0.6771              &  0.6986 \\ \hline

\color{black} Clustering & 0.7101          & 0.7160             & 0.6413              & 0.6891   \\ \hline

ERM pruning & 0.7806          & 0.7601             & 0.6538              & 0.7315   \\ \hline
KD          & 0.7241          & 0.7554             & 0.6385              & 0.706    \\ \hline
SparseTrain & 0.7619          & 0.7899             & 0.6571              & 0.7363   \\ \hline

HHISS       & 0.8153          & 0.8024             & 0.6983              & 0.772    \\ \hline
\end{tabular}
}
        \caption{  Accuracy for Models Trained with Post-dose Data}
        \label{tab:postdose}
    \end{minipage}
\end{table}

\subsubsection{Q3: Can the model effectively handle unseen stressors?}\label{rq3}

Evaluating generalizability to unseen stressors is essential for expanding and evaluating the model's OOD applicability across diverse domains and scenarios. 
However, our collected datasets evaluated earlier were collected using the same protocol and thus involve identical stressors, as discussed in Section \ref{self-dataset-discussion}.

\begin{figure}
    \centering
    
    \includegraphics[width =0.5\linewidth]{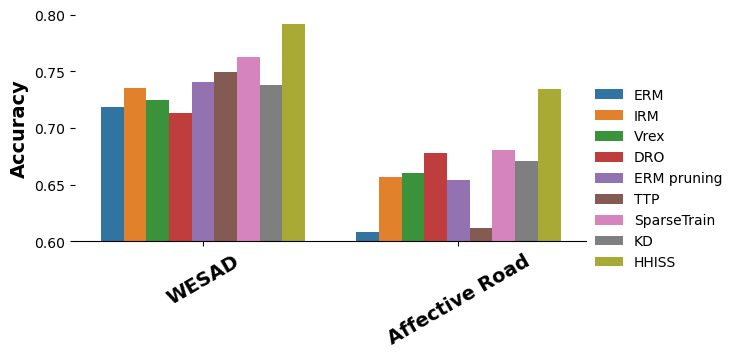}

    \caption{Evaluate on off-the-shelf dataset when train on Control dataset}
\vskip -3ex
    \label{fig:offtheshelf}
 
\end{figure}

To evaluate OOD robustness with unseen stressors, we tested our model trained on the Control dataset against two off-the-shelf stress detection datasets, WESAD and AffectiveROAD. These datasets feature different data collection setups and stress-inducing factors. Figure \ref{fig:offtheshelf} illustrates the evaluation accuracy on the WESAD and AffectiveROAD datasets, with the model trained on the Control-training dataset using different approaches. On both datasets, the HHISS model outperforms the baselines, demonstrating its superior OOD robustness.

Notably, among the two datasets, all models demonstrate superior OOD performance on the WESAD dataset. This is because the WESAD dataset was collected in a controlled lab environment and includes some stressors (e.g., arithmetic tasks) similar to those used in the protocol for collecting our control dataset, as discussed in Section \ref{self-dataset-discussion}, which the models were trained on.

In contrast, the AffectiveROAD dataset was collected during real driving sessions, involving entirely different stressors. Despite this significant distribution shift, HHISS achieves an OOD accuracy of approximately 74\%, highlighting its practical effectiveness.




\subsubsection{Q4: Is HHISS specifically tailored to be trained on our dataset, or can it be effectively trained on other datasets as well?}\label{rq4}

This section examines whether HHISS maintains superior OOD robustness when trained on off-the-shelf datasets. Successfully doing so would demonstrate that its ability to learn invariant stress features is not limited to our collected dataset, highlighting its versatility and reliability for stress detection across diverse real-world scenarios and populations.

 \begin{table}[]
\centering
\resizebox{0.7\linewidth}{!}{
\begin{tabular}{l|r|rrr}
\hline
            & \multicolumn{1}{l|}{In distribution}                                 & \multicolumn{3}{l}{Out of distribution}                             \\ \hline
Dataset     & \multicolumn{1}{l|}{WESAD}        & \multicolumn{1}{l}{Control}      & \multicolumn{1}{l}{Predose}      & \multicolumn{1}{l}{Postdose}     \\ \hline
ERM         & 0.8771                           & 0.6911                            & 0.5306                           & 0.5563                           \\ \hline
IRM         & 0.878                            & 0.7258                            & 0.5638                           & 0.619                            \\ \hline
DRO         & 0.8624                           & 0.7299                            & 0.5569                           & 0.6424                           \\ \hline
Vrex        & 0.8625                           & 0.7271                            & 0.5866                           & 0.6411                           \\ \hline
TTP &0.7292&0.6806&0.5018&0.5833 \\ \hline

\color{black} DANN        &     0.7816                             & 0.701                                  &    0.566                              &        0.6068                          \\ \hline
\color{black} MLDG        &     0.7682                             & 0.7106                                  &    0.5486                              &        0.6154                          \\ \hline
\color{black} Clustering  & 0.7529                        &  0.7113                           &   0.5434                          & 0.6021                             \\ \hline
ERM pruning  & 0.8593                           & 0.7649                            & 0.5864                           & 0.676                            \\ \hline
KD          & 0.8847                           & 0.7217                            & 0.5165                           & 0.6132                           \\ \hline
SparseTrain & 0.8753                           & 0.7261                            & 0.5524                           & 0.6461                           \\ \hline
HHISS       & 0.864                            & 0.8204                            & 0.6324                           & 0.7023                           \\ \hline
\end{tabular}
}
\caption{Balanced accuracy of models trained on WESAD datasets.}
\vskip -5ex
\label{table:WESAD}
\end{table}

\textcolor{black}{Table \ref{table:WESAD} compares HHISS to baseline methods when trained on the WESAD dataset, following the evaluation setup discussed in Section \ref{dataset_setup}. WESAD dataset features less versatile stressors, consisting of only two stress-inducing tasks, including public speaking—a task not included in our data collection protocol, as noted in Section \ref{self-dataset-discussion}. The limited variety of stressors makes model training simpler by requiring fewer stress patterns to be learned, leading to higher in-distribution performance on the WESAD test set across all methods. However, when tested on out-of-distribution (OOD) data, performance drops for all approaches due to limited exposure to diverse conditions. Still, HHISS consistently outperforms all baselines, demonstrating strong robustness and generalization. These results highlight the need for training on datasets with more diverse stressors to improve OOD performance and show that HHISS remains effective even when trained on external data.}


\subsubsection{\textcolor{black}{Q5: Performance analysis of HHISS trained on datasets with different distributions.}}\label{rq5}

\textcolor{black}{This section examines whether HHISS maintains strong OOD robustness when trained jointly on two distinct datasets with distributional heterogeneity. While training solely on self-collected healthy data may lead to a better fit due to its consistency, we also evaluated the HHISS model’s performance when trained on both the WESAD and Control datasets. These datasets exhibit distributional differences, as discussed in Section \ref{lmm_public_data}, and are evaluated using the setup described in Section \ref{dataset_setup}. As shown in Table \ref{table:WESADcontrol}, HHISS outperforms all baseline methods across all test distributions. Compared to results in Tables \ref{table:baseline} and \ref{table:WESAD}, its in-distribution performance improves slightly when trained on the combined dataset. Notably, all baseline methods also show enhanced performance on the Control test set under joint training. However, performance on the WESAD test set declines slightly, likely due to WESAD’s underrepresentation in the combined training set (12 WESAD subjects vs. 36 Control subjects).}

\textcolor{black}{Importantly, when evaluated on the out-of-distribution Pre-dose and Post-dose OUD test sets, HHISS performs slightly worse than when trained solely on the Control dataset (see Table \ref{table:baseline}). This indicates that merging datasets with different distributions, like our Control set and WESAD, does not enhance OOD performance. Instead, it poses challenges for domain generalization methods, including HHISS, in learning invariant feature representations across mismatched distributions of training instances, potentially limiting generalization to unseen distributions.}

 \begin{table}[]
\centering
\resizebox{0.7\linewidth}{!}{\color{black} 
\begin{tabular}{l|ll|ll}
\hline
            & \multicolumn{2}{l|}{In distribution}                                 & \multicolumn{2}{l}{Out of distribution}                             \\ \hline
Dataset     & \multicolumn{1}{l}{WESAD}        & \multicolumn{1}{l|}{Control}      & \multicolumn{1}{l}{Predose}      & \multicolumn{1}{l}{Postdose}     \\ \hline
ERM         & 0.7368                           & 0.8018                            & 0.608                            & 0.6691                           \\ \hline
IRM         & 0.7002                           & 0.7756                            & 0.6221                           & 0.6236                           \\ \hline
DRO         & 0.7048                           & 0.818                             & 0.6119                           & 0.6711                           \\ \hline
Vrex        & 0.741                           & 0.7927                            & 0.613                           & 0.226                           \\ \hline
TTP  &0.6635&0.8124&0.6044&0.6819 \\ \hline
\color{black} DANN  &0.7459&0.7455&0.5989&0.6565 \\ \hline
\color{black} MLDG        &     0.7714                             & 0.7692                                  &    0.5865                              &        0.631                          \\ \hline
ERM Pruning  & 0.7425                           & 0.8266                            & 0.6258                           & 0.6695                           \\ \hline

\color{black} Clustering  & 0.6913                        &  0.8115                           &   0.5726                          & 0.6232                             \\ \hline
KD          & 0.7564                           & 0.7849                            & 0.6123                           & 0.7077                           \\ \hline
SparseTrain & 0.7614                           & 0.7979                            & 0.6251                            & 0.7253                           \\ \hline
HHISS       & 0.8212                           & 0.8566                            & 0.6911                           & 0.7708                           \\ \hline
\end{tabular}
}
\caption{Balanced accuracy of models trained on the combined WESAD and Control datasets.}
\vskip -5ex
\label{table:WESADcontrol}
\end{table}

\subsubsection{\textcolor{black}{Q6: Can Models Trained in the lab Settings Generalize to In-the-Wild field Scenarios?}}\label{rq6}
\textcolor{black}{In real-world settings, stress often arises from a complex mix of social, cultural, and psychological factors, which are difficult to replicate in controlled environments. Prior evaluations in Section \ref{rq5} using datasets like WESAD and AffectiveRoad primarily focus on stress induced by isolated stressors, such as those induced by controlled laboratory settings or driving tasks. To better assess model performance and generalizability under in-the-wild field conditions, we evaluated our HHISS model trained on Control data on two real-world field datasets: the Nurse Work Stress dataset (Section \ref{nurse-dataset} ) and our self-collected OUD Daily Stress dataset (Section \ref{OUD-field-dataset} ). Table \ref{table:in-the-wild} presents the evaluation result of the HHISS model and other domain generalization baselines in these in-the-wild field settings. Consistently, HHISS achieves the best performance on both datasets, highlighting its robustness and ability to generalize under complex, in-the-wild field conditions.}

\par
\begin{figure}[h!]
    \centering
    \includegraphics[scale=0.48]{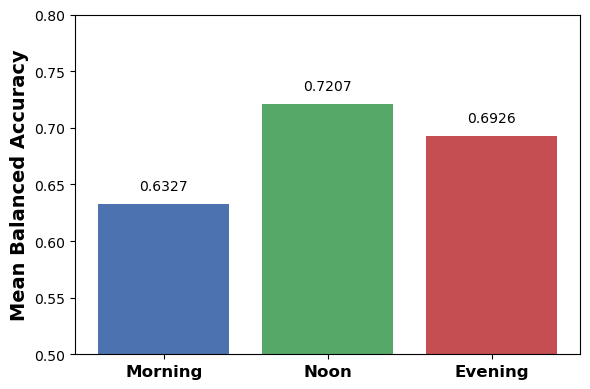} 
    \caption{\textcolor{black}{Balanced accuracy of HHISS model across different survey time windows}}
    \label{fig:survey_window}
    \vskip -3ex
\end{figure}

 
\textcolor{black}{Notably, the OUD Daily Stress dataset includes five days of physiological data from two OUD individuals, with stress self-reports collected three times per day. As shown in Figure \ref{fig:ema}, participants reported more stress events in the evening compared to the morning. Since for these participants, methadone is typically administered in the morning (Section \ref{participant-description-OUD}), we hypothesize that its effects may suppress stress responses earlier in the day, which then diminish over time, peaking by evening. This dataset allows us to assess intra-OUD-individual generalizability across different times of day. As shown in Table \ref{fig:survey_window}, the HHISS model maintains consistent performance, achieving mean balanced accuracies of 63.27\% in the morning, 72.07\% at noon, and 69.26\% in the evening, demonstrating its ability to generalize within OUD individuals across time in real-world, in-the-wild field settings. Notably, performance was lowest in the evening, potentially due to greater physiological variability associated with higher stress and vulnerability reported by the OUD participants at that time, which may have introduced the highest distribution shifts that challenged the model.}



 \begin{table}[]
\centering
\resizebox{0.4\linewidth}{!}{\color{black}
\begin{tabular}{l|llll}
\hline
            & \multicolumn{2}{l}{Nurse Stress} & \multicolumn{2}{l}{OUD Daily stress} \\ \hline
            & Acc            & Macro F1        & Acc             & Macro F1          \\ \hline
ERM         & 0.5704         & 0.5565          & 0.5157          & 0.5004            \\ \hline
IRM         & 0.5882          & 0.5881          & 0.5279           & 0.5277            \\ \hline
DRO         & 0.5849         & 0.5652          & 0.5666           & 0.5088            \\ \hline
Vrex        & 0.5773          & 0.5529          & 0.5557          & 0.4974            \\ \hline
\color{black} DANN        &         0.5779       &   0.5771              &    0.6003             &  0.5924                 \\ \hline
MLDG        &         0.0.5743       &   0.0.5741              &    0.6038            &  0.6027                \\ \hline

Clustering  & 0.5856         &   0.5761        &     0.5512            &    0.5004              \\ \hline
ERM pruning& 0.5938         & 0.5751          & 0.6202          & 0.6133            \\ \hline
KD          &   0.5728             &    0.5604             & 0.5642          & 0.5048            \\ \hline
SparseTrain &   0.5831             &    0.5697             & 0.6258          & 0.5927            \\ \hline
HHISS       & 0.6178         & 0.6141          & 0.7274          & 0.7231            \\ \hline
\end{tabular}
}
\caption{Models performance under in-the-wild field setup.}
\vskip -5ex
\label{table:in-the-wild}
\end{table}

\subsection{Ablation study}\label{results-ablation}
This section discusses a comprehensive ablation by modifying different components of HHISS, pruning amount during training, and network complexity. All evaluations in this section follow the dataset setup defined in Section \ref{experiment-setup}, with the model trained on the Control-training set and evaluated on the Control-testing, Pre-dose, and Post-dose sets.
\subsubsection{Ablation on Approach Components}\label{ablation-approach-componenets}
\begin{figure}[h!]
    \centering
    \includegraphics[scale=0.48]{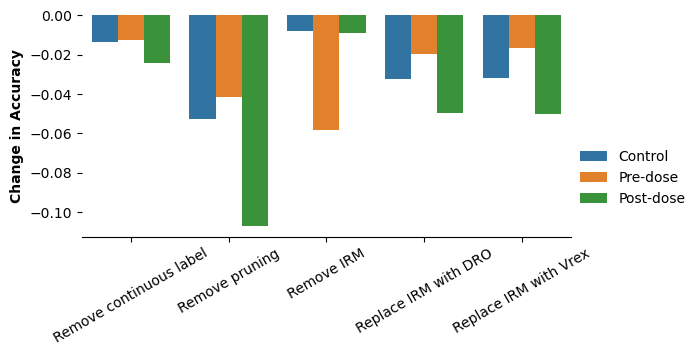} 
    \caption{Accuracy variations relative to HHISS when different components are removed.}
    \label{fig:ablation}
    \vskip -3ex
\end{figure}

Figure \ref{fig:ablation} shows the ablation evaluation, where different components of HHISS are removed or replaced while keeping other components unchanged. Performance drops with the removal of any component. 

\textcolor{black}{Notably, the first step of HHISS, as outlined in Section \ref {Algo:IRM-training-1}, is to develop an embedding generation network through IRM regularization that enables learning causally invariant feature extraction for stress detection across all individuals from the training set (considering them as domains). Two of our baselines, DRO and Vrex, also incorporate regularization to enable the extraction of robust features/embeddings, but our ablation study showed that replacing IRM with either method led to inferior results. DRO regularization encourages the models to learn embeddings whose worst-case performance across domains (here individuals) is optimized, typically by focusing on domains with higher loss, often leading to aligned embeddings that reduce maximum risk. However, it does not explicitly promote causal invariance. Vrex regularization encourages the model to learn embeddings/features that yield consistent empirical risk across training domains (individuals in the training set) by penalizing the variance in losses. While it focuses on risk stability rather than feature alignment, V-REx also lacks a causal perspective. In stress detection, physiological signals often shift significantly across domains, such as between control and pre- or post-dose OUD groups. Thus, leveraging causal invariance—as IRM does—enables better generalization across these heterogeneous conditions.}

Moreover, ablation or replacement of the pruning component causes the largest decline. This underscores the key role of subject-wise pruning in improving generalizability, while other components also enhance overall effectiveness.

\subsubsection{Ablation on Pruning Amount}
\begin{figure}[h]
    \centering
    \begin{minipage}{0.48\textwidth}
        \centering
        \includegraphics[width =0.88\linewidth]{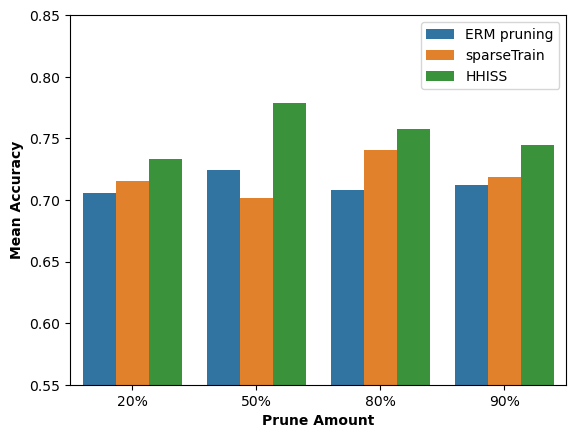} 
        \caption{Performance comparison of HHISS at different prune amounts (PA).}
        \label{fig:radar_chart_prune_amount}
    \end{minipage}
    \hfill
    \begin{minipage}{0.48\textwidth}
        \centering
        \includegraphics[width =0.88\linewidth]{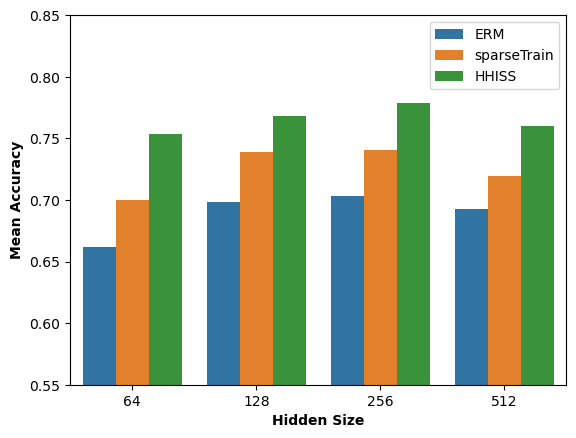} 
        \caption{Performance comparison of HHISS for different hidden size (HS).}
        \label{fig:hidden_size}
    \end{minipage}
\end{figure}

Figure \ref{fig:radar_chart_prune_amount} illustrates the impact of various pruning levels by presenting the mean accuracy achieved across Control-testing, Pre-dose, and Post-dose datasets. Both ERM pruning and HHISS achieved the best performance at 50\% pruning, while SparseTrain performed best at 80\% pruning. The pruning amount is a hyperparameter that needs to be tuned for each approach to achieve optimal performance. Pruning introduces a trade-off: slight pruning preserves more information, while aggressive pruning focuses on retaining only invariant features \cite{zhou2022sparse}. However, excessive pruning can remove too much information, including invariant features, leading to performance degradation. Notably, our approach outperformed the baselines across all pruning levels, emphasizing the effectiveness of HHISS.

\begin{figure}[htbp]
    \centering

\end{figure}

\vskip -3ex
\subsubsection{Ablation on Network Complexity}

During our evaluation, the neural network consists of four hidden layers of the same size. Figure \ref{fig:hidden_size} compares different hidden layer sizes using ERM, SparseTrain, and HHISS by presenting the mean accuracy achieved across Control-testing, Pre-dose, and Post-dose datasets. All approaches achieve best performance with a size of 256.  Larger models capture more abstract features, benefiting datasets similar to the training set but hurting generalizability in datasets with larger distribution shifts, like Pre-dose and Post-dose. Overall, HHISS outperforms SparseTrain and ERM across all model sizes.

\subsubsection{\textcolor{black}{Ablation on backbone Network}}\label{backbone-abltation}
\textcolor{black}{This section explores how different backbone architectures affect HHISS and our baseline domain generalization approaches. Since the input features are non-sequential, we evaluate two alternative backbones: a ResNet architecture with 1D convolutional layers, and a multi-head attention mechanism applied along the feature dimension.  The results, presented in Tables \ref{tab:resnet} and \ref{tab:mhd}, show that compared to DNN backbone results shown in Table \ref{table:baseline}, by utilizing other backbones, all baselines achieve improved performance in the in-distribution setting. However, they exhibit reduced performance in the out-of-distribution setting, particularly when using MHA as the backbone. This decline may be attributed to overfitting due to the higher complexity of these architectures. In contrast, our HHISS method consistently outperforms all baselines across both in-distribution and out-of-distribution scenarios with these backbones. These results suggest that enhancing the backbone can improve generalization within known distributions, but may not extend to entirely unseen distributions. Lastly, we emphasize that our HHISS method is compatible with different model architectures and consistently delivers strong performance. }

\begin{table}[h!]
    \centering
    \begin{minipage}{0.48\linewidth}
        \centering
        \resizebox{\linewidth}{!}{ \color{black}
        \begin{tabular}{l|l|ll}
\hline
            & \multicolumn{1}{l|}{In distribution} & \multicolumn{2}{l}{Out of distribution}                            \\ \hline
Dataset    & \multicolumn{1}{l|}{Control}     & \multicolumn{1}{l}{Predose} & \multicolumn{1}{l}{Postdose} \\ \hline
ERM         & 0.7824                               & 0.6545                          & 0.7166                           \\ \hline
IRM         & 0.7959                               & 0.6736                          & 0.7152                           \\ \hline
DRO         & 0.7948               & 0.6805            & 0.7041          \\ \hline
Vrex        & 0.8065                               & 0.6456                          & 0.7287                           \\ \hline

ERM pruning & 0.8086                               & 0.6671                          & 0.7272                           \\ \hline
KD          & 0.802                               & 0.676                         & 0.727                           \\ \hline
SparseTrain & 0.7851                & 0.6794            & 0.7413            \\ \hline

HHISS       & 0.8165                               & 0.7445                           & 0.7684                           \\ \hline
\end{tabular}
        }
        \caption{Balanced Accuracy with ResNet as the Backbone}
        \label{tab:resnet}
    \end{minipage}%
    \hfill
    \begin{minipage}{0.48\linewidth}
        \centering
        \resizebox{\linewidth}{!}{ \color{black}
        \begin{tabular}{l|l|ll}
\hline
            & \multicolumn{1}{l|}{In distribution} & \multicolumn{2}{l}{Out of distribution}                            \\ \hline
Dataset    & \multicolumn{1}{l|}{Control}     & \multicolumn{1}{l}{Predose} & \multicolumn{1}{l}{Postdose} \\ \hline
ERM         & 0.8063                               & 0.584                           & 0.5536                           \\ \hline
IRM         & 0.8029                               & 0.609                          & 0.6114                           \\ \hline
DRO         & 0.7984                & 0.6277           & 0.5794            \\ \hline
Vrex        & 0.7917                               & 0.5935                          & 0.5728                           \\ \hline

ERM pruning & 0.8141                               & 0.6086                          & 0.6007                           \\ \hline
KD         & 0.783                                     &           0.5866                        &   0.5427                               \\ \hline
SparseTrain & 0.7916                & 0.5895            & 0.647             \\ \hline

HHISS       & 0.7922                               & 0.6349                          & 0.6708                          \\ \hline
\end{tabular}
}
        \caption{Balanced Accuracy with MHA as the Backbone}
        \label{tab:mhd}
    \end{minipage}
\end{table}



\subsubsection{Remove one of modalities} \label{ablation_modality}
\begin{figure}
    \centering
    \includegraphics[width=\linewidth]{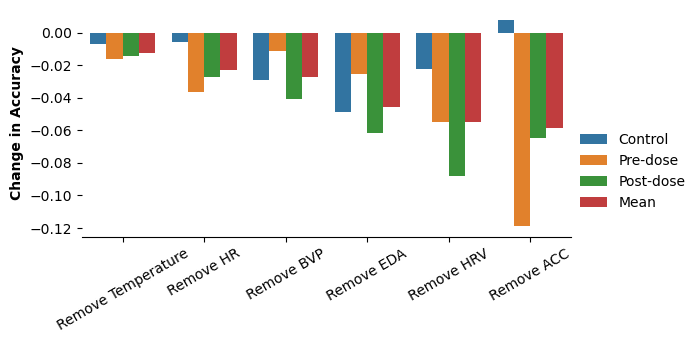}

    \caption{Accuracy variations relative to HHISS when different input modality are removed.}
    \label{fig:Modality_ablation}
 \vskip -3ex
\end{figure}

This section examines the impact of excluding specific input modalities on HHISS's performance, with detailed results presented in figure \ref{fig:Modality_ablation}. While removing skin temperature, heart rate, and BVP results in a slight performance decline, excluding EDA, HRV, and accelerometer data leads to a significant degradation in performance. This observation aligns with existing research, as EDA and HRV are considered reliable indicators of stress response \cite{ghiasi2020assessing,szakonyi2021efficient}. Interestingly, excluding the accelerometer modality leads to mixed results: a slight performance improvement on the Control dataset but a significant performance drop in the Pre-dose and Post-dose datasets. This suggests that the accelerometer enhances the model's generalizability. Its removal appears to result in overfitting to the Control group's data. This phenomenon may stem from differences in physiological responses between the OUD and Control groups, while their gestures and movements under stress and calm conditions show similarities. These findings underscore the variability in the dataset and the necessity of incorporating all modalities to achieve generalizable outcomes.

\subsection{Design Choices' Empirical Justification}\label{results-empirical-justification}

This section empirically discusses the design choice of HHISS to facilitate effective OOD generalizability.

\subsubsection{Use task as domain}

\begin{figure}[h]
    \centering
    \begin{minipage}{0.51\textwidth}
        \centering
        \includegraphics[width =0.83\linewidth]{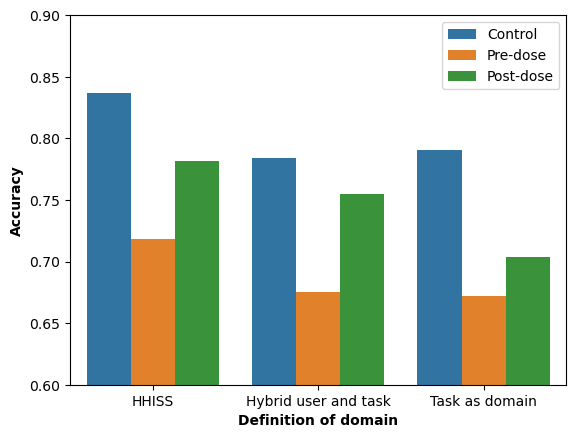}
        \caption{Comparison using different domain definitions.}
        \label{fig:DefinationOfDomain}
    \end{minipage}
    \hfill
    \begin{minipage}{0.48\textwidth}
        \centering
        \includegraphics[width =0.88\linewidth]{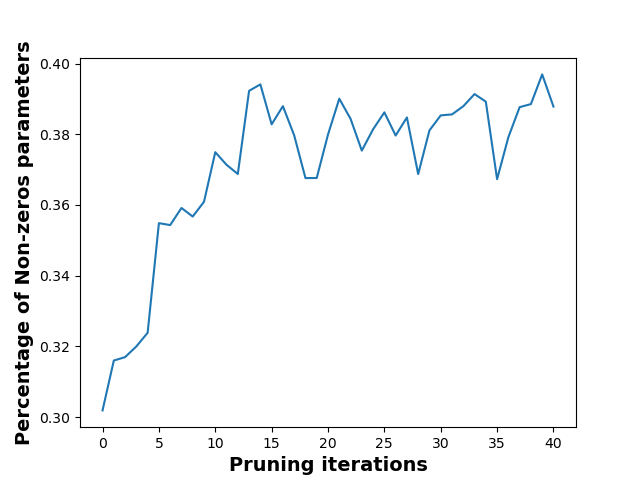}
        \caption{Intersectional `Mask' size across iterations.}
        \label{fig:importnace}
    \end{minipage}
    \vskip -3ex
\end{figure}

In HHISS, as outlined in Section \ref{approach-intersection-pruning}, each user is treated as a unique domain. An alternative is to define the tasks or stressors as domains and extract pruned subnetworks across tasks. Figure \ref{fig:DefinationOfDomain} shows the accuracy changes for both `task-wise' and `combined user- and task-wise hybrid' intersections. Our user-wise approach performs best across all variations. While different tasks elicit varied psychophysiological responses, they are limited in number and balanced with sufficient samples. However, significant heterogeneity exists among individuals within the same task, which task-invariant feature extraction fails to capture. Interestingly, the combined user- and task-wise approach also performs poorly. This is likely due to the insufficient number of samples per-individual-and-par-task, making it challenging to learn an OOD-robust invariant representation.




\subsubsection{`Sparse-to-sparse' training enabling consistent intersectional `Mask'}
\label{Sparse-to-sparse}

In Section \ref{our-approach-DST}, HHISS employs `sparse-to-sparse' training to ensure subject-wise pruned networks converge, forming a consistent intersection `Mask' for effective performance. Figure \ref{fig:importnace} shows the percentage of non-zero parameters after intersecting pruned masks at each iteration. We apply 50\% pruning per subject, leaving about 50\% of their parameters non-zero. After the intersection, initially, 30\% of parameters remain, increasing to 40\% as training progresses. This indicates that the networks converge, creating a stable intersection `Mask' that retains enough information for generalizable, effective modeling.

\subsubsection{Generalizability through Loss Landscapes Flatness}\label{loss-landscape-perspective}
\begin{figure}
    \centering
    \begin{subfigure}[h]{.3\linewidth}
    \includegraphics[width =\linewidth]{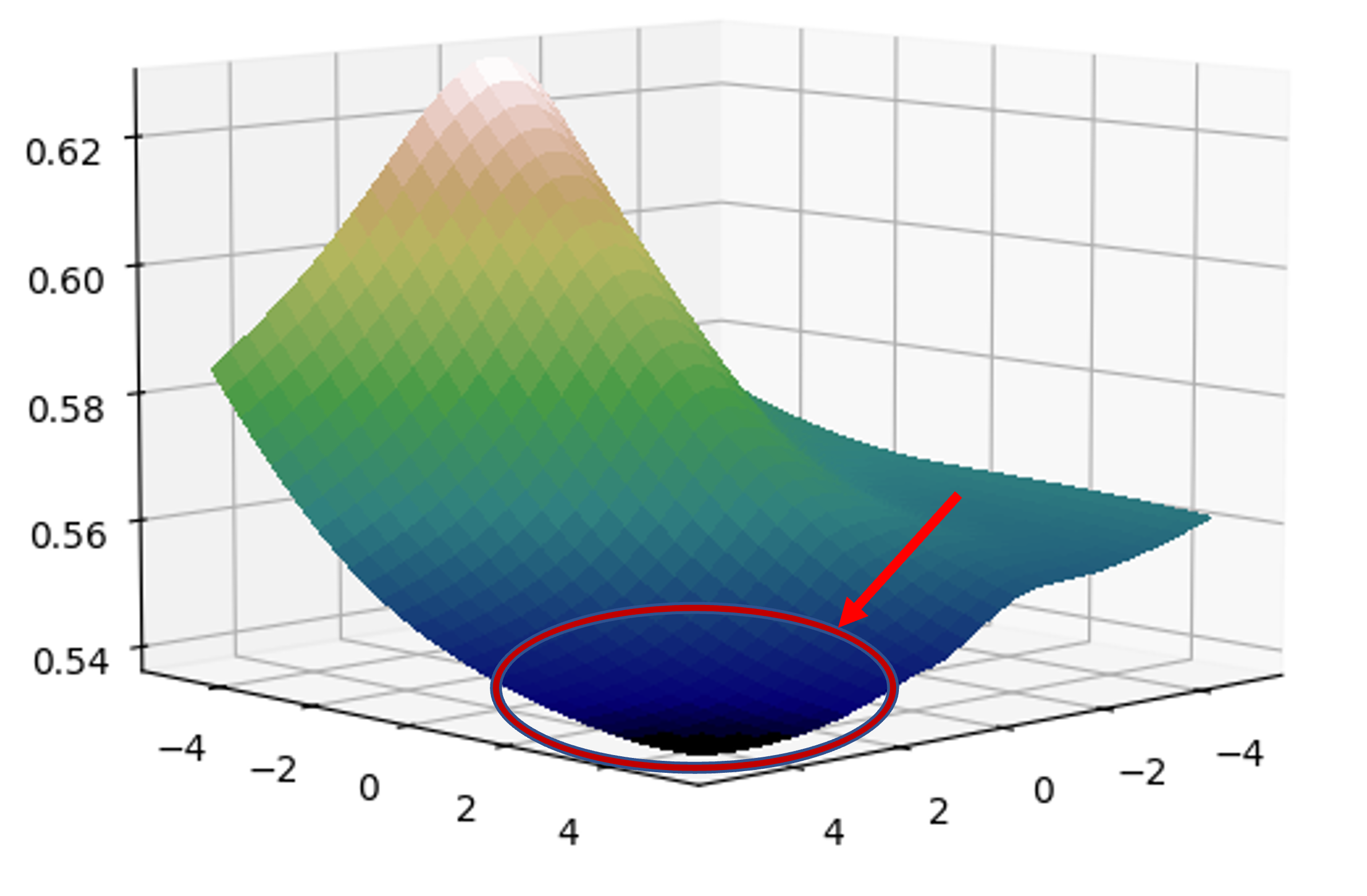}
        \caption{HHISS}
    \end{subfigure}
    \begin{subfigure}[h]{.3\linewidth}
    \includegraphics[width =\linewidth]{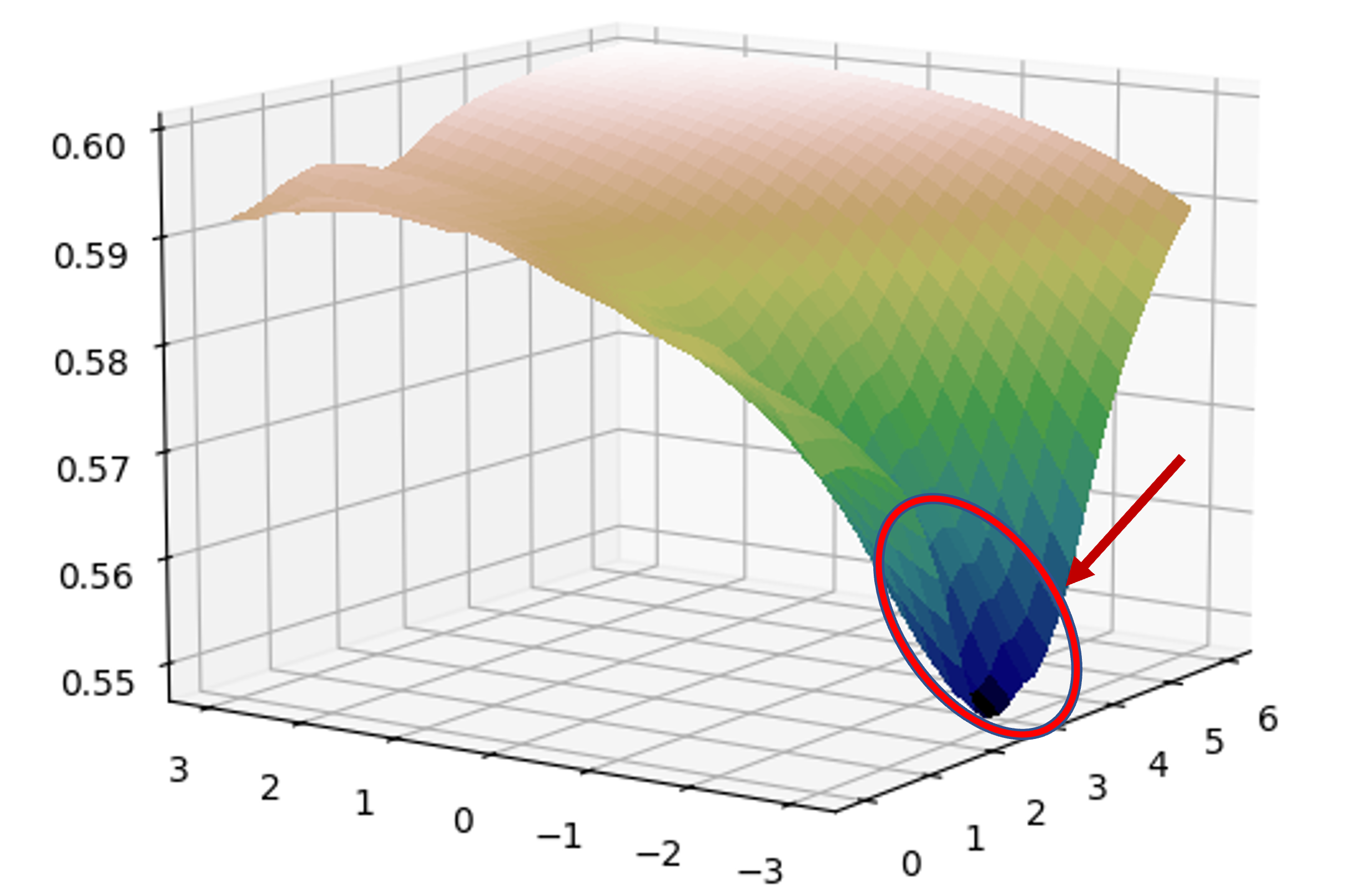}
    \caption{ERM}
    \end{subfigure}
    \begin{subfigure}[h]{.3\linewidth}
    \includegraphics[width =\linewidth]{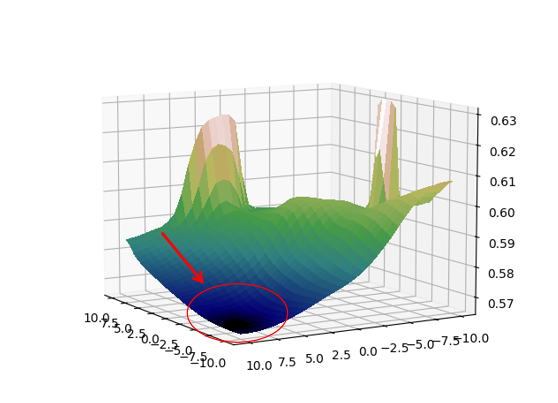}
    \caption{SparseTrain}
    \end{subfigure}
    
    \caption{Losslandscape Comparison}
\vskip -3ex
    \label{fig:loss_landscape}
 
\end{figure}

In Section \ref{our-approach-KD}, HHISS uses continuous labels to improve generalizability by converging on a flatter loss landscape. Figure \ref{fig:loss_landscape} shows a 3D visualization of the loss landscape, where the XY plane represents the parameter subspace and the z-axis shows loss values. Consistent with prior studies \cite{li2018visualizing,cha2021swad,rangamani2020loss,zhang2023generalization,wang2021embracing}, which suggest that flatter loss landscapes improve generalization, HHISS achieves a flatter minima (dark black region, marked by the red circle) compared to the baselines.

\begin{figure}
    \centering
    \includegraphics[width=\linewidth]{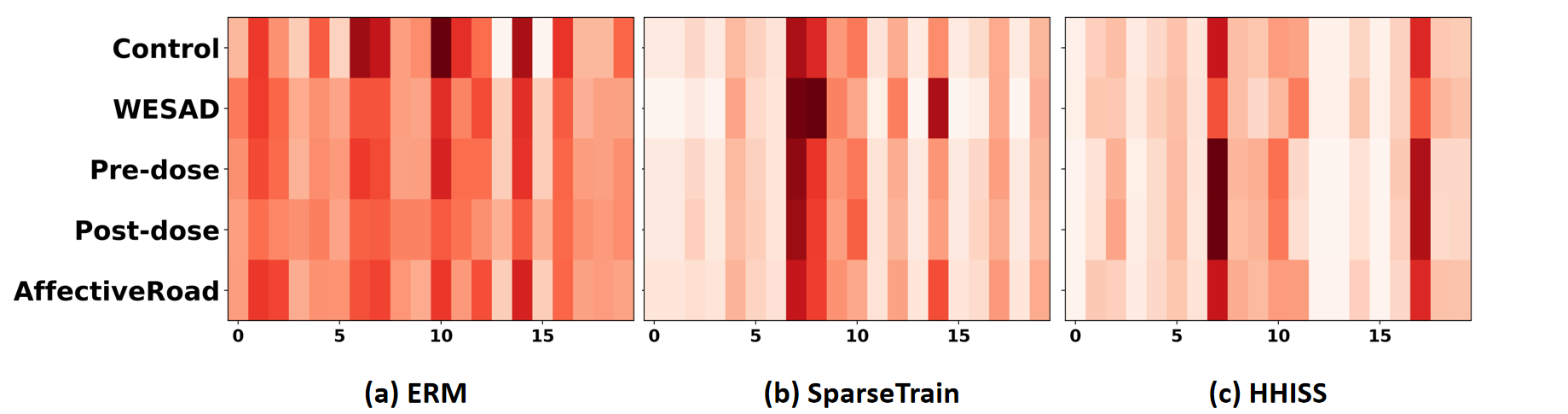}
    
    %
    

    \caption{Saliency Maps of Selected Features Across Datasets for different approaches.}
    \label{fig:loss_gradient}
 \vskip -3ex
\end{figure}

\subsubsection{Distribution-invariant Feature Extraction}
This section highlights HHISS's ability to identify distribution-invariant features for effective OOD robustness. Figure \ref{fig:loss_gradient} displays saliency maps \cite{simonyan2013deep} for randomly selected 20 features. These maps reveal which input features most influence the output (through darker color), with larger absolute gradients indicating stronger impact \cite{simonyan2013deep}. HHISS consistently produces similar saliency patterns across datasets, whereas baseline methods generate more variable and noisier results. This suggests that the ERM approach extracts different features depending on the dataset, increasing the risk of overfitting. In contrast, HHISS consistently focuses on the same features across datasets, demonstrating its effectiveness in identifying invariant features.

Notably, both SparseTrain and HHISS assign high saliency scores to features 7–11 and 17, which correspond to Accelerometer L2 5th percentile, Accelerometer Z minimum, Accelerometer L2 minimum, EDA phasic L2 minimum, EDA tonic L2 sum, and HRV MCVNN. These influential features originate from the Accelerometer, EDA, and HRV modalities. This aligns with our observations from the modality ablation study in Section \ref{ablation_modality}, where removing EDA, HRV, and Accelerometer resulted in the most significant performance drop. This further confirms that these modalities are crucial for accurate feature prediction.


\subsection{Embedding visualization}
\label{distribution-shif-validation}

\begin{figure}
    \centering
    \begin{subfigure}[h!]{0.45\linewidth}
    \includegraphics[width=\linewidth]{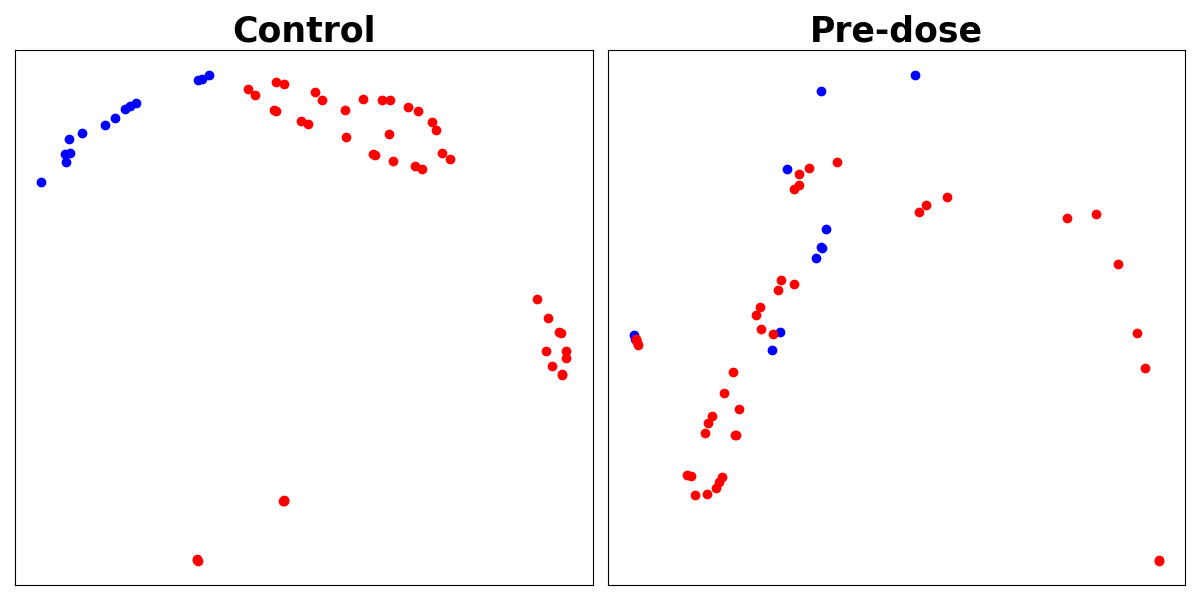}
       \caption{ERM}
    \end{subfigure}%
    \begin{subfigure}[h]{0.45\linewidth}
    \includegraphics[width=\linewidth]{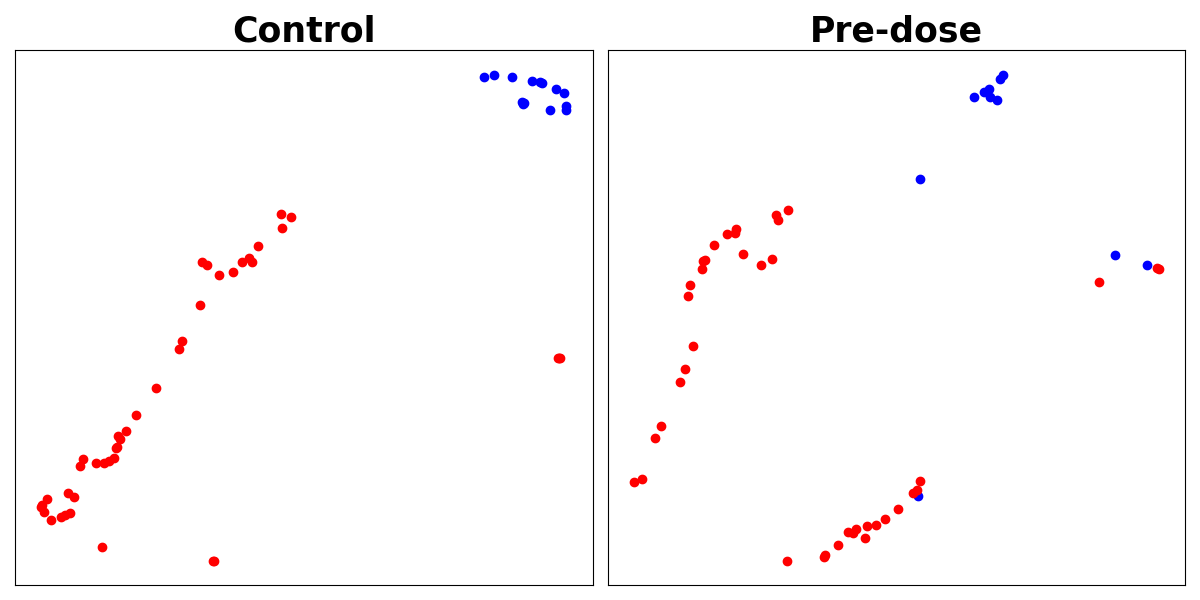}
    \caption{HHISS}
    \end{subfigure}
    
    \caption{Embedding distribution of intermediate layers when trained with Control dataset.}{(black points indicate calm data, red points represent stress data)}
    \label{fig:embedding}

\end{figure}

To better understand the features extracted by different approaches, we analyze the intermediate embeddings of two models trained on the control-training set using the HHISS and ERM approaches.
We utilize T-SNE to reduce the dimensionality to two for visualization purposes. Figure \ref{fig:embedding} presents the embeddings of data from a control subject and a Pre-dose subject, extracted from intermediate layers of the ERM and HHISS approach. black points indicate calm data, while red points represent stress data. The embeddings derived from the ERM method effectively differentiate the control group data but struggle to distinguish the Pre-dose group data. In contrast, our HHISS approach successfully differentiates both the control and Pre-dose group data, demonstrating its OOD effectiveness.

\subsection{Subject-wise variability of each group}
\label{group-variability}
\begin{figure}
    \centering
    \includegraphics[width=0.8\linewidth]{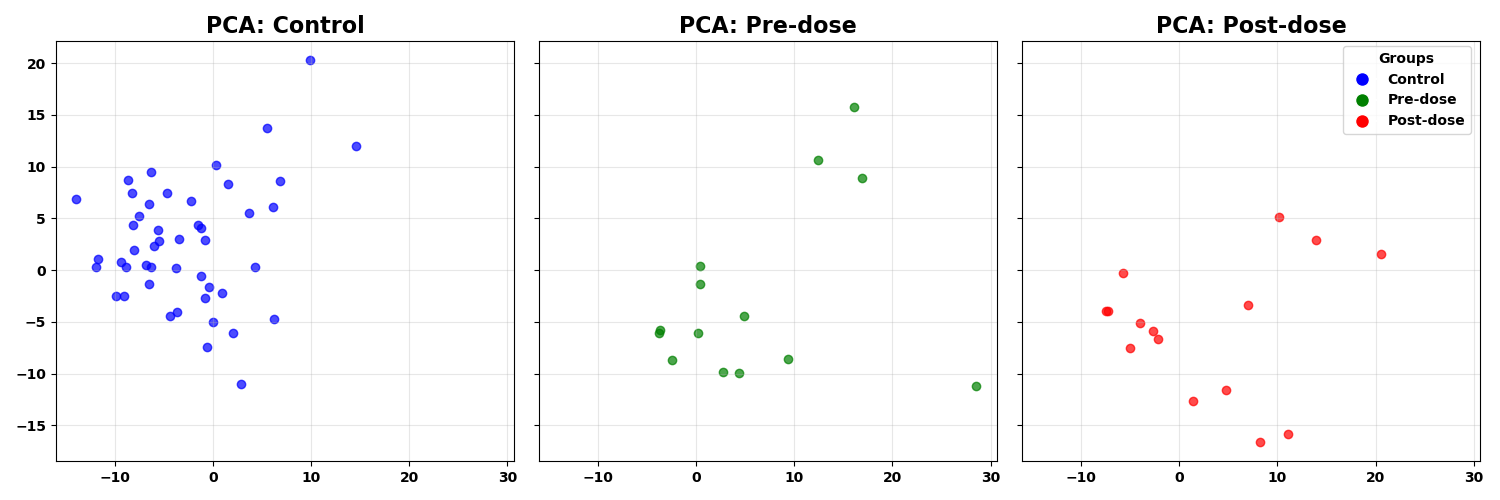}
    \caption{Subject wise PCA visualization of raw features}
    \label{fig:variability}
\vskip -3ex
\end{figure}

To investigate why the model trained with the control group achieves the best overall generalizability, we visualize the subject-wise variability for each group. First, we calculate the mean of each subject across all features and then apply Principal Component Analysis (PCA) to reduce the dimensionality to two for visualization. PCA is effective at preserving variability while reducing dimensions. As shown in Figure \ref{fig:variability}, each data point represents a subject. We observe that the control group exhibits higher variability compared to the Pre-dose and Post-dose datasets. This can be attributed to two factors: on the one hand, the physiological response of stress within the Pre-dose and Post-dose datasets can be more homogeneous than control. However, on the other hand, the control group includes a larger number of subjects, which naturally introduces more variability. Our HHISS approach is able to capture and leverage this inter-subject variability to learn invariant features. The larger inter-subject variability in the control group benefits the learning process, which is why HHISS achieves better OOD generalization when trained on control group data.

\section{Discussion}

This section discusses 
HHISS's scalability on mobile and edge platforms, limitations, and future research directions.



\subsection{Training and Inference Scalability Across Deployment Platforms}

\begin{table}[h!]
\centering
\resizebox{0.7\linewidth}{!}{
\begin{tabular}{|c|c|c|c|c|}
\hline
\textbf{Platform} & \textbf{Approach} & \multicolumn{1}{|p{3cm}|}{\centering \textbf{Training Time (s)}} & \multicolumn{1}{|p{3cm}|}{\centering \textbf{Process (RSS) \ Memory (MBs)}} & \multicolumn{1}{|p{3cm}|}{\centering \textbf{Resource Utilization (\%)}} \\ \hline
\multirow{4}{3cm}{\centering AMD Ryzen 9 5950X 16-Core Processor} 
 & IRM & 1.77 & 1383 & 11.98 \\ \cline{2-5} 
 & KD & 0.89 & 1383 & 5.26 \\ \cline{2-5} 
 & HHISS & 2.47 & 1387 & 17.77 \\ \hline
\multirow{4}{2.5cm}{\centering Apple M1} 
 & IRM & 1.27 & 189.75 & 0.14 \\ \cline{2-5} 
 & KD & 0.28 & 240.08 & 0.125 \\ \cline{2-5} 
 & HHISS & 2.34 & 240.4 & 0.15 \\ \hline

\multirow{4}{2.5cm}{\centering Raspberry Pi 5} 
 & IRM & 3.2 & 265.89 & 42 \\ \cline{2-5} 
 & KD & 1.51 & 264.91 & 45 \\ \cline{2-5} 
 & HHISS & 7.89 & 268.91 & 45.75 \\ \hline
\multirow{4}{3cm}{\centering AMD Ryzen 9 7950X 16-Core Processor} 
 & IRM & 0.44 & 484.15 & 23.9 \\ \cline{2-5} 
 & KD & 0.08 & 482.86 & 24.41 \\ \cline{2-5} 
 & HHISS & 0.89 & 499.02 & 24.9 \\ \hline

 \multirow{3}{3cm}{\centering Google Pixel 6} 
 & IRM & 1.35 & 264.41 & 49 \\ \cline{2-5} 
 & KD & 0.25 & 262.71 & 48 \\ \cline{2-5} 
 & HHISS & 11.201 & 265.64 & 49 \\ \hline
\end{tabular}
}
\caption{Training Time Scalability Analysis with Resource Utilization}
\label{table:training_platform_comparison}
\vskip -3ex
\end{table}

\begin{table}[ht]
\centering
\resizebox{0.7\linewidth}{!}{
\begin{tabular}{|c|c|c|c|c|}
\hline
\textbf{Platform} & \textbf{Approach} & \multicolumn{1}{|p{3cm}|}{\centering \textbf{Inference Time (s)}} & \multicolumn{1}{|p{3cm}|}{\centering \textbf{Process (RSS) \ Memory (MBs)}} & \multicolumn{1}{|p{3cm}|}{\centering \textbf{Resource Utilization (\%)}} \\ \hline
\multirow{4}{3cm}{\centering AMD Ryzen 9 5950X 16-Core Processor} 
 & IRM & 0.001 & 468.98 & 17 \\ \cline{2-5}
 & KD & 0.001 & 471.18 & 17.5 \\ \cline{2-5}
 & HHISS & 0.001 & 470.4 & 18 \\ \hline
\multirow{4}{2.5cm}{\centering Apple M1} 
 & IRM & 0.001 & 187.83 & 0.175 \\ \cline{2-5}
 & KD & 0.001 & 238.22 & 0.175 \\ \cline{2-5}
 & HHISS & 0.001 & 238.47 & 0.175 \\ \hline

\multirow{4}{2.5cm}{\centering Raspberry Pi 5} 
 & IRM & 0.009 & 260.36 & 53 \\ \cline{2-5}
 & KD & 0.008 & 260.39 & 45 \\ \cline{2-5}
 & HHISS & 0.008 & 260.41 & 45 \\ \hline
\multirow{4}{3cm}{\centering AMD Ryzen 9 7950X 16-Core Processor} 
 & IRM & 0.001 & 474.68 & 17.5 \\ \cline{2-5}
 & KD & 0.001 & 474.39 & 17 \\ \cline{2-5}
 & HHISS & 0.001 & 486.57 & 17.5 \\ \hline

\multirow{4}{3cm}{\centering Google Pixel 6} 
 & IRM & 0.001 & 90 & 0.01 \\ \cline{2-5}
 & KD & 0.001 & 97.5 & 0.01 \\ \cline{2-5}
 & HHISS & 0.001 & 100.9 & 0.01 \\ \hline
\end{tabular}
}
\caption{Inference Time Scalability Analysis with Resource Utilization}
\label{table:inference_platform_comparison}
\vskip -5ex
\end{table}

Tables \ref{table:training_platform_comparison} and \ref{table:inference_platform_comparison} compare training and inference performance of HHISS and baselines: IRM and KD,
on a wide range of platforms, including resource-constrained devices like Raspberry Pi 5, and Google Pixel 6, as well as high-performance systems like the AMD Ryzen 9 5950X, AMD Ryzen 9 7950X, and Apple M1. Metrics include training/inference time (s), process RSS memory (MBs), and average resource utilization (\%), reflecting computational efficiency and scalability. These metrics align with methodologies in prior works \cite{jaiswal2024tinystressnet,rachakonda2019stress,xu2023practically}.

\textit{During the training phase:} HHISS demonstrated strong performance across platforms, taking 7.89 seconds on the Raspberry Pi 5, with memory usage of 268.91 MB, showcasing its suitability for low-power edge devices. On high-performance platforms like the AMD Ryzen 9 7950X, HHISS exhibited training times as low as 0.89 seconds. On the Google Pixel 6, HHISS achieved a training time of 11.20 seconds, with memory usage of 265.64 MB and resource utilization of 49\%, further demonstrating its adaptability across diverse devices.

\textit{During the inference phase:} Inference times were significantly lower across all platforms. On the Raspberry Pi 5, inference was completed in 0.008 seconds with minimal memory overhead. Mobile devices like the Google Pixel 6 demonstrated exceptional efficiency, with inference times of 0.001 seconds and negligible resource utilization. On high-performance platforms like the AMD Ryzen 9 7950X and 5950X, inference times remained consistent at 0.01 seconds, respectively, with resource utilization around 17.5\%.

Overall, HHISS shows excellent scalability and efficiency, achieving competitive runtimes with minimal power and memory usage across edge, mobile, and high-performance platforms. This makes it an ideal candidate for both edge-based and cloud-offloaded applications, further reinforcing its versatility and practicality \cite{liu2022energy}.

\subsection{\textcolor{black}{Feature Section for OOD generalizability}\label{feature-select-discussion}}

\begin{figure}[h!]
    \centering
    \includegraphics[scale=0.38]{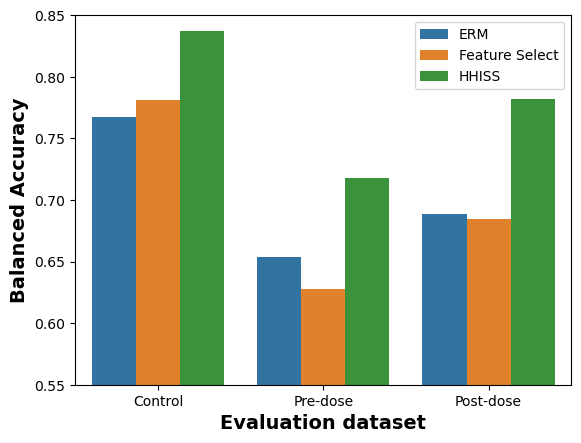}
    \caption{Balanced accuracy on ERM, HHISS, and feature select.}
    \label{fig:feature_select}
\end{figure}

\textcolor{black}{Our analysis in Section \ref{motivation} identifies $257$ features that are consistently non-significant across dataset-wise group comparisons, having the potential of having distributional robustness.  To evaluate this, we trained an ERM-based model using only these features. However, as shown in Figure \ref{fig:feature_select}, when trained solely on the control dataset, the model exhibited poor generalization under distribution shifts in the pre- and post-dose OUD groups. Such a result may stem from the limitation that removing potentially informative-but variable—features in pursuit of statistical invariance may inadvertently discard important discriminative attributes/information. This is in line with prior works \cite{ng2008feature} which report that simple feature selection strategies ignore non-linear correlation between the input and output, ultimately impairing model performance in OOD settings. This highlights the need for an effective strategy to learn embedding-level representations that are robust to domain shifts while retaining predictive relevance.}

\subsection{Limitations and Future Direction}\label{limitation-future-direction}
The limitations of HHISS and future research directions are outlined below.

\begin{enumerate}
    \item HHISS uses subject-wise gradient pruning and intersection, adding some computational overhead during training, though the training time across platforms remains comparable to OOD baselines. While HHISS applies unstructured pruning to identify equitable invariant features, unstructured pruning does not reduce runtime on generic platforms \cite{sadou2022inference}, making HHISS's execution time similar to that of the baselines.
    \item Despite advancements in the OOD generalization on the opioid use disorder group, a performance gap persists between the Pre-and Post-dose OUD group. The achievement of OOD fairness in performance across different distribution shifts remains a critical area for future research and development.
    \item As discussed in Section \ref{experiment-setup}, our network architectures are based on those benchmarked in the literature. We also explored self-attention and multi-head attention networks but found no performance improvement. We hypothesize this is due to limited training data and the simplicity of wearable stress signals, leading to overfitting in more complex networks and compromising OOD generalizability.
    \item This paper focuses on wearable-based stress sensing due to its practical relevance, growing popularity\cite{vos2023generalizable,gonzalez2023wearables}, and use in the health domain. However, human sensing heterogeneity extends to many other applications, and future work could apply our framework to different modalities and use cases.
    \item \textcolor{black}{The OUD dataset used in this study is limited in size, with data collected from 25 unique individuals in a controlled lab setting and 2 additional individuals in the in-the-wild field setting. This reflects a well-documented challenge in recruiting participants with substance use disorders, stemming from factors such as stigma, co-occurring mental health conditions, irregular participation, high dropout rates, and ethical and safety concerns \cite{de2012challenges,anderson2018ethical,allan2019ethics}. Prior relevant studies involving substance abuse disorder populations have reported similar recruitment difficulties and small sample sizes \cite{gullapalli2019body,gullapalli2021opitrack}.}
    
    \item While our OUD dataset is unique, it has limited condition variations related to medication use. However, stress responses in OUD individuals may vary due to other factors (Section \ref{realted-work-OUD}), and \textcolor{black}{a longitudinal study across a larger population} is needed to investigate OOD robustness across these variations, which is a future research direction.
    \item \textcolor{black}{Our approach, HHISS, does not enhance robustness across different physiological sensor types, modalities, and devices. Even sensor variability for the same modality across different devices, such as differences in sampling rates, signal quality, and device-specific preprocessing, can affect model performance \cite{bent2020investigating}. For example, the Microsoft Band 2 has an accelerometer sampling rate of 8Hz, rather than the Empatica E4's 32Hz; such a variability may have an effect on activity or stress detection efficacy across these devices. The HHISS approaches or our baselines do not incorporate mechanisms in their design choices to address such cross-device or cross-modality (i.e., sensor types) variability, which falls outside the scope of this study and can be a potential future work.}
    
\end{enumerate}




\section{Conclusion}
This paper pioneers OOD robust, wearable-based stress sensing, trained on a healthy population but generalizing to various previously unknown distributions, including health population groups like individuals with opioid use disorder (OUD). By extracting equitable, distribution-shift invariant features, HHISS sets the foundation for more \textit{equitable} and \textit{inclusive} stress sensing, applicable even to other health conditions or populations with scarce data. This enhances the accessibility and effectiveness of stress management tools across diverse contexts. Comprehensive evaluations across five different stress distributions confirm HHISS's feasibility and scalability as a practical mobile stress-sensing solution.

\bibliographystyle{ACM-Reference-Format}
\bibliography{sample-base}

\appendix

\section{Methodological Transparency \& Reproducibility Appendix}
To ensure reproducibility, we provide detail about the backbone model used for training and hyperparameters used for Algorithm \ref{alg:1}, Equation \ref{eq:finetune} in Appendix \ref{appendix:Architecture}. Additionally, Appendix \ref{appendix:code} provides details of the source code.

\textcolor{black}{\subsection{Model Architecture and Training Configuration}}
\label{appendix:Architecture}

We use a 4-layer DNN as the main classification backbone. In the ablation study, we explore ResNet and multi-head self-attention (MHA) as alternative architectures. Model details are provided below.

The 4-layer DNN takes input vectors of shape (batch size, 340) and processes them through a sequence of fully connected layers with ReLU activations and dropout for regularization. 
\begin{table}[H]
\centering
\scriptsize
\resizebox{0.7\textwidth}{!}{%
\begin{tabular}{|l|l|l|l|l|}
\hline
Layer & Type            & Input Size & Output Size & Detail          \\ \hline
fc1   & Linear          & 340        & 128         & ReLU Activation \\ \hline
fc2   & Linear          & 128        & 128         & ReLU Activation           \\ \hline
fc3   & Linear          & 128        & 128         & ReLU Activation          \\ \hline
fc4   & Linear (output) & 128        & num class  & Softmax Activation        \\ \hline
\end{tabular}
}
\caption{DNN backbone details}
\label{tab:DNN_backbone}
\end{table}

 The ResNet backbone is a 1D convolutional neural networks with residual connections and a fully connected classifier for sequence classification. It processes inputs of shape (batch size, 340) by first applying a 1D convolution (channel-inflation to 64), followed by two residual blocks that enhance gradient flow and feature reuse. A max-pooling layer reduces the sequence length by half before flattening and passing the output through a sequence of fully connected layers. The model ends with a softmax activation for class probabilities.

 \begin{table}[H]
\centering
\scriptsize
\resizebox{0.7\textwidth}{!}{%
\begin{tabular}{|l|l|l|l|l|}
\hline
Component     & Type          & Input Size & Output Size  & Details                            \\ \hline
initial\_conv & Conv1d        & (1, 340)   & (64, 340)    & Kernel=3, Padding=1                \\ \hline
res\_block1   & ResidualBlock & (64, 340)  & (64, 340)    & 2 × Conv1d + ReLU + Dropout        \\ \hline
res\_block2   & ResidualBlock & (64, 340)  & (64, 340)    & 2 × Conv1d + ReLU + Dropout        \\ \hline
pool          & MaxPool1d     & (64, 340)  & (64, 170)    & Kernel=2                           \\ \hline
flatten       & Reshape       & (64, 170)  & 10880        & 64 × 170                           \\ \hline
output\_proj  & MLP + Softmax & 10880      & num classes & 3-layer MLP + final classification \\ \hline
\end{tabular}
}
\caption{ResNet backbone details}
\label{tab:resnet_backbone}
\end{table}

 The MHA backbone is a feature-wise multi-head self-attention mechanism. Each of the 340 input features is treated as an individual token, allowing the model to learn contextual interactions between features. It uses learned query, key, and value projections and applies scaled dot-product attention over feature dimensions. After computing attention across multiple heads, the outputs are aggregated and passed through a fully connected projection head for classification.

 \begin{table}[H]
\centering
\scriptsize
\resizebox{0.7\textwidth}{!}{%
\begin{tabular}{|l|l|l|l|l|}
\hline
Component      & Type                & Input Shape      & Output Shape      & Details                            \\ \hline
input\_proj    & Linear              & (B, 340, 1)      & (B, 340, 128)     & Projects scalar to embedding dim   \\ \hline
qkv\_proj      & Linear              & (B, 340, 128)    & (B, 340, 384)     & Combined Q, K, V projection        \\ \hline
split\_heads   & Reshape + Transpose & (B, 340, 128)    & (B, 4, 340, 32)   & 4 heads, 32-dim each               \\ \hline
attention      & Scaled Dot Product  & (B, 4, 340, 340) & (B, 4, 340, 32)   & Softmax over similarity matrix     \\ \hline
combine\_heads & Reshape             & (B, 4, 340, 32)  & (B, 340, 128)     & Concatenate head outputs           \\ \hline
attended.mean  & Reduce (mean)       & (B, 340, 128)    & (B, 128)          & Feature-wise average               \\ \hline
output\_proj   & MLP + Softmax       & (B, 128)         & (B, num classes) & 3-layer MLP + final classification \\ \hline
\end{tabular}
}
\caption{MHA backbone details}
\label{tab:mha_backbone}
\end{table}

For all evaluations, we use consistent training configurations to ensure fair comparisons in the ablation study. The Adam optimizer is used with a fixed learning rate of 0.0001, and cross-entropy loss serves as the objective for binary stress detection. A dropout rate of 0.1 is applied across layers to mitigate overfitting. Input features are standardized to a length of 340, and hidden dimensions are set to 128 based on the architecture. For Algorithm \ref{alg:1}, the sparse-to-sparse threshold (T) is set to 70\%, the pruning amount (K) to 50\%, and the number of training rounds (R) to 50. In Equation \ref{eq:finetune}, the weighting factor for the continuous label ($\lambda$) is 0.5. The weighting factor for IRM regularization ($\beta$) is 0.3.

The optimization goal of HHISS method can be formulate as :

\begin{equation}
\begin{aligned}
    Loss = CrossEntropy(y,\hat{y})+ \beta \sum_{S_i \in S} \Vert \nabla_{w} R^{e} (W,\Phi)  \Vert  + \lambda CrossEntropy(\hat{y}_{o},\hat{y})
\end{aligned}
\label{eq:finetune}
\end{equation}

Here, $y$ denotes the ground-truth labels, and $\hat{y}$ is the model's output logits. The loss function consists of three components: The first term is the supervised learning loss using cross-entropy. The second term is an IRM regularization term adopted from \cite{arjovsky2019invariant}, weighted by a factor $\beta$ and $S$ representing all subjects in the training set. The third term introduces consistency with continuous labels, where $\hat{y}_{o}$ is the output of the  over-parameterized model, weighted by $\lambda$.

This loss function is used both for training the model and for computing importance scores during pruning. To determine parameter importance, we compute the loss on user data, backpropagate to obtain gradients for each layer, and then calculate the importance score by multiplying the gradient with the corresponding weight value. The bottom K\% of parameters with the lowest importance scores are then pruned by setting them to zero.

\subsection{Code}
\label{appendix:code}
 The code is provided at \href{https://anonymous.4open.science/r/HHISS_IMWUT-B665/README.md}{anonymous link}. It includes the pre-trained HHISS model checkpoint, source code for training both the IRM-regularized over-parameterized and HHISS models, feature extraction scripts, \textcolor{black}{stats analysis code and results using a Linear Mixed Effects Model(LMM),} and instructions for running the experiments. Seed values used for evaluation are specified within the source files. 

\textcolor{black}{\section{ WESAD Training with complex backbone network architecture}}
State-of-the-art work (SOTA) has reported over 90\% (in-distribution) accuracy on the WESAD dataset \cite{choi2022attention}. To evaluate the performance on this benchmark, we apply our domain generalization method HHISS to an SOTA model designed for WESAD \cite{choi2022attention}. This model uses raw PPG signals as input and employs an LSTM with an attention mechanism for stress detection. To better assess the model's generalizability, we deviate from the original paper's evaluation setup, which used only one subject for testing, by following the evaluation setup discussed in Section \ref{dataset_setup}, which uses three subjects in the test set and the remaining subjects for training. As shown in Table~\ref{table:rawppg}, our domain generalization methods, including HHISS, improve performance in the in-distribution setting (in WESAD testset). Specifically, our HHISS method boosts balanced accuracy by over 5\% compared to normal ERM training. However, under the out-of-distribution setting, all methods perform poorly. This is expected, as the original SOTA model was not designed for domain generalization and only uses the PPG modality. Notably, even in this evaluation, HHISS outperforms all baselines. These results demonstrate that our HHISS approach can be applied across different model architectures and highlight the importance of using multi-modal inputs (all of our other evaluations used multimodal input), and possibly simpler network architecture for effective out-of-distribution generalization.

 \begin{table}[]
\centering
\resizebox{0.7\linewidth}{!}{ 
\begin{tabular}{l|r|rrr}
\hline
            & \multicolumn{1}{l|}{In distribution}                     & \multicolumn{3}{l}{Out of distribution}                    \\ \hline
Dataset     & \multicolumn{1}{l|}{WESAD} & \multicolumn{1}{l}{Control} & \multicolumn{1}{l}{Predose} & \multicolumn{1}{l}{Postdose} \\ \hline
ERM         & 0.8866                    & 0.4735                       & 0.6477                      & 0.5621                       \\ \hline
IRM         & 0.8976                    & 0.5354                       & 0.6718                      & 0.5514                       \\ \hline
DRO        &0.8707&0.5307&0.6436&0.5451        \\ \hline
Vrex        & 0.8993                    & 0.5537                       & 0.6367                      & 0.5532                       \\ \hline
TTP &0.9121&0.4700&0.6402&0.5540 \\ \hline

ERM pruning  & 0.9184                    & 0.5192                       & 0.6552                      & 0.5419                       \\ \hline
KD          & 0.9272                    & 0.5404                       & 0.6609                      & 0.5336                       \\ \hline
SparseTrain &         0.8423                  & 0.5285                             &    0.6562                         &     0.537                         \\ \hline
HHISS       & 0.9331                    & 0.5454                       & 0.6669                      & 0.5791                       \\ \hline
\end{tabular}
}
\caption{Applying Domain Generalization Methods to the SOTA Model on the WESAD Dataset}
\vskip -5ex
\label{table:rawppg}
\end{table}

\textcolor{black}{\section{Detailed results for LMM evaluations}}

\label{LMM_results_appendix}
This section presents the detailed LLM results for each pair-wise evaluation: Control vs Predose (Table \ref{tab:lmm_detailed_control_predose}), Control vs Postdose (Table \ref{tab:lmm_detailed_control_postdose}), Predose vs Postdose (Table \ref{tab:lmm_detailed_predose_postdose}), Control vs AffectiveRoad (Table \ref{tab:lmm_detailed_control_affectiveroad}),Control vs WESAD (Table \ref{tab:lmm_detailed_control_wesad}), WESAD vs AffectiveRoad (Table \ref{tab:lmm_detailed_wesad_affectiveroad}). 

Each table presents the features which showed statistical significance ($p<0.05$) consistently across both the stress and calm task along with the coefficient estimates, standard errors, and $95\%$ confidence intervals $[CI\ Low-CI\ High]$. 
\subsection{Control vs Pre-Dose}
\begin{table}[H]
\centering
\scriptsize
\resizebox{0.7\textwidth}{!}{%
\begin{tabular}{|c|c|c|c|c|c|c|}
\hline
\textbf{Feature} & \textbf{Condition} & \textbf{Coef.} & \textbf{Std. Err.} & \textbf{95\% CI Low} & \textbf{95\% CI High} & \textbf{p-value} \\
\hline
\multirow{2}{*}{HR L2 Kurtosis} & Calm & 2.504 & 3.127 & 0.935 & 4.074 & 0.002 \\ \cline{2-7}
& Stress & 9.283 & 2.337 & 4.702 & 13.863 & 0.0000 \\ \hline
\multirow{2}{*}{HR Kurtosis} & Calm & 2.845 & 1.148 & 0.594 & 5.096 & 0.0132 \\ \cline{2-7}
& Stress & 6.477 & 1.426 & 3.682 & 9.272 & 0.0000 \\ \hline
\multirow{2}{*}{HR n below mean} & Calm & 5.172 & 1.380 & 2.467 & 7.877 & 0.0002 \\ \cline{2-7}
& Stress & 5.118 & 1.634 & 1.915 & 8.321 & 0.0017 \\ \hline
\multirow{2}{*}{HR Peaks} & Calm & 3.853 & 1.890 & 0.150 & 7.557 & 0.0414 \\ \cline{2-7}
& Stress & 12.642 & 5.131 & 2.586 & 22.698 & 0.0137 \\ \hline
\multirow{2}{*}{HR Skewness} & Calm & 1.225 & 0.571 & 0.106 & 2.344 & 0.0319 \\ \cline{2-7}
& Stress & 9.586 & 3.395 & 2.932 & 16.240 & 0.0048 \\ \hline
\multirow{2}{*}{HR L2 Skewness} & Calm & 1.377 & 0.631 & 0.141 & 2.613 & 0.0290 \\ \cline{2-7}
& Stress & 5.215 & 1.739 & 1.806 & 8.623 & 0.0027 \\ \hline
\multirow{2}{*}{Phasic EDA L2 Root Mean Square} & Calm & 2.594 & 1.009 & 0.618 & 4.571 & 0.0101 \\ \cline{2-7}
& Stress & 3.438 & 1.491 & 0.515 & 6.362 & 0.0211 \\ \hline
\multirow{2}{*}{Phasic EDA Kurtosis} & Calm & -5.505 & 2.135 & -9.689 & -1.321 & 0.0099 \\ \cline{2-7}
& Stress & -9.262 & 3.604 & -16.327 & -2.198 & 0.0102 \\ \hline
\multirow{2}{*}{Skin Temp Kurtosis} & Calm & 4.393 & 1.521 & 1.411 & 7.375 & 0.0039 \\ \cline{2-7}
& Stress & 9.079 & 1.918 & 5.319 & 12.838 & 0.0000 \\ \hline
\multirow{2}{*}{Skin Temp Min} & Calm & 2.014 & 0.908 & 0.234 & 3.795 & 0.0266 \\ \cline{2-7}
& Stress & 3.295 & 0.964 & 1.406 & 5.184 & 0.0006 \\ \hline
\multirow{2}{*}{Skin Temp n below mean} & Calm & 3.313 & 1.376 & 0.615 & 6.010 & 0.0161 \\ \cline{2-7}
& Stress & 4.557 & 1.307 & 1.996 & 7.117 & 0.0005 \\ \hline
\multirow{2}{*}{Skin Temp Skewness} & Calm & 4.379 & 1.297 & 1.837 & 6.921 & 0.0007 \\ \cline{2-7}
& Stress & 6.944 & 3.303 & 0.470 & 13.418 & 0.0355 \\ \hline
\multirow{2}{*}{Skin Temp L2 Lineintegral} & Calm & 67.715 & 31.701 & 5.581 & 129.848 & 0.0327 \\ \cline{2-7}
& Stress & 138.485 & 63.309 & 14.401 & 262.569 & 0.0287 \\ \hline
\multirow{2}{*}{Skin Temp L2 peaks} & Calm & 11.990 & 3.824 & 4.494 & 19.485 & 0.0017 \\ \cline{2-7}
& Stress & 21.888 & 9.808 & 2.664 & 41.113 & 0.0256 \\ \hline
\multirow{2}{*}{Tonic EDA L2 Kurtosis} & Calm & -3.862 & 1.624 & -7.044 & -0.680 & 0.0174 \\ \cline{2-7}
& Stress & -5.951 & 2.271 & -10.403 & -1.500 & 0.0088 \\ \hline
\multirow{2}{*}{Tonic EDA L2 95th Percentile} & Calm & 0.308 & 0.090 & 0.132 & 0.485 & 0.0006 \\ \cline{2-7}
& Stress & 0.558 & 0.209 & 0.150 & 0.967 & 0.0074 \\ \hline
\multirow{2}{*}{Tonic EDA Entropy} & Calm & 0.218 & 0.092 & 0.038 & 0.398 & 0.0178 \\ \cline{2-7}
& Stress & 0.507 & 0.215 & 0.087 & 0.927 & 0.0181 \\ \hline
\end{tabular}}
\caption{LMM results for common significant features across stress and calm tasks (Control vs Predose)}
\label{tab:lmm_detailed_control_predose}
\end{table}
\subsection{Control vs Post-Dose}

\begin{table}[H]
\centering
\scriptsize
\resizebox{0.7\textwidth}{!}{%
\begin{tabular}{|c|c|c|c|c|c|c|}
\hline
\textbf{Feature} & \textbf{Condition} & \textbf{Coef.} & \textbf{Std. Err.} & \textbf{95\% CI Low} & \textbf{95\% CI High} & \textbf{p-value} \\
\hline
\multirow{2}{*}{HR n below mean} & Calm & 6.187 & 1.390 & 3.463 & 8.910 & 0.0000 \\ \cline{2-7}
& Stress & 6.005 & 1.860 & 2.361 & 9.650 & 0.0001 \\ \hline
\multirow{2}{*}{BVP Energy} & Calm & 6.735 & 2.216 & 2.393 & 11.078 & 0.0024 \\ \cline{2-7}
& Stress & 5.408 & 2.701 & 0.114 & 10.702 & 0.0453 \\ \hline
\multirow{2}{*}{BVP Sum} & Calm & -6.735 & 2.216 & -11.078 & -2.393 & 0.0024 \\ \cline{2-7}
& Stress & -5.408 & 2.701 & -10.702 & -0.114 & 0.0453 \\ \hline
\multirow{2}{*}{Skin Temp min} & Calm & 2.497 & 0.922 & 0.691 & 4.304 & 0.0067 \\ \cline{2-7}
& Stress & 3.932 & 1.179 & 1.621 & 6.242 & 0.0009 \\ \hline
\multirow{2}{*}{Skin Temp n below mean} & Calm & 3.608 & 1.373 & 0.917 & 6.298 & 0.0086 \\ \cline{2-7}
& Stress & 4.291 & 1.291 & 1.760 & 6.822 & 0.0009 \\ \hline
\end{tabular}}
\caption{LMM results for common significant features across calm and stress tasks (Control vs Postdose)}
\label{tab:lmm_detailed_control_postdose}
\end{table}

\subsection{Pre-Dose vs Post-Dose}

\begin{table}[H]
\centering
\scriptsize
\resizebox{0.7\textwidth}{!}{%
\begin{tabular}{|c|c|c|c|c|c|c|}
\hline
\textbf{Feature} & \textbf{Condition} & \textbf{Coef.} & \textbf{Std. Err.} & \textbf{95\% CI Low} & \textbf{95\% CI High} & \textbf{p-value} \\
\hline
\multirow{2}{*}{Phasic EDA peaks} & Calm & -2.646 & 1.093 & -4.788 & -0.503 & 0.0155 \\ \cline{2-7}
 & Stress & -2.607 & 0.994 & -4.555 & -0.659 & 0.0087 \\
\hline
\end{tabular}}
\caption{LMM results for common significant features across stress and calm tasks (Predose vs Postdose)}
\label{tab:lmm_detailed_predose_postdose}
\end{table}
\subsection{Control vs AffectiveRoad}

\begin{table}[h!]
\centering
\scriptsize
\resizebox{0.7\textwidth}{!}{%
\begin{tabular}{|c|c|c|c|c|c|c|}
\hline
\textbf{Feature} & \textbf{Condition} & \textbf{Coef.} & \textbf{Std. Err.} & \textbf{95\% CI Low} & \textbf{95\% CI High} & \textbf{p-value} \\
\hline
\multirow{2}{*}{ACC L2 Lineintegral} & Calm & 27.257 & 4.379 & 18.675 & 35.839 & 0.0000 \\ \cline{2-7}
& Stress & 40.891 & 13.940 & 13.569 & 68.212 & 0.003 \\ \hline
\multirow{2}{*}{ACC y 95th Percentile} & Calm & 25.130 & 5.175 & 14.987 & 35.273 & 0.0000 \\ \cline{2-7}
& Stress & 29.348 & 13.523 & 2.845 & 55.852 & 0.0300 \\ \hline
\multirow{2}{*}{ACC z Lineintegral} & Calm & 27.257 & 4.379 & 18.675 & 35.839 & 0.0000 \\ \cline{2-7}
& Stress & 40.891 & 13.940 & 13.569 & 68.212 & 0.0034 \\ \hline
\multirow{2}{*}{ACC z 95th Percentile} & Calm & 25.130 & 5.175 & 14.987 & 35.273 & 0.0000 \\ \cline{2-7}
& Stress & 29.348 & 13.523 & 2.845 & 55.852 & 0.0300 \\ \hline
\multirow{2}{*}{BVP Energy} & Calm & 2.853 & 1.455 & 0.001 & 5.705 & 0.0499 \\ \cline{2-7}
& Stress & 16.073 & 4.349 & 7.549 & 24.597 & 0.0002 \\ \hline
\multirow{2}{*}{BVP Skewness} & Calm & 2.894 & 1.284 & 0.376 & 5.411 & 0.0243 \\ \cline{2-7}
& Stress & 12.981 & 4.509 & 4.144 & 21.818 & 0.0040 \\ \hline
\multirow{2}{*}{BVP Sum} & Calm & -2.853 & 1.455 & -5.705 & -0.001 & 0.0499 \\ \cline{2-7}
& Stress & -16.073 & 4.349 & -24.597 & -7.549 & 0.0002 \\ \hline
\multirow{2}{*}{HR L2 skewness} & Calm & 8.510 & 2.246 & 4.108 & 12.912 & 0.0002 \\ \cline{2-7}
& Stress & 32.158 & 6.491 & 19.437 & 44.879 & 0.0000 \\ \hline
\multirow{2}{*}{Skin Temp n below mean} & Calm & -7.745 & 3.766 & -15.126 & -0.364 & 0.0397 \\ \cline{2-7}
& Stress & 6.046 & 1.512 & 3.083 & 9.010 & 0.0001 \\ \hline
\end{tabular}
}
\caption{LMM results for common significant features across stress and calm task (Control vs AffectiveRoad)}
\label{tab:lmm_detailed_control_affectiveroad}
\end{table}
\newpage
\subsection{Control vs WESAD}

\begin{table}[H]
\centering
\scriptsize
\resizebox{0.65\textwidth}{!}{%
\begin{tabular}{|c|c|c|c|c|c|c|}
\hline
\textbf{Feature} & \textbf{Condition} & \textbf{Coef.} & \textbf{Std. Err.} & \textbf{95\% CI Low} & \textbf{95\% CI High} & \textbf{p-value} \\
\hline
\multirow{2}{*}{ACC y Interquartile Range} & Calm & 22.815 & 7.246 & 8.614 & 37.016 & 0.002 \\ \cline{2-7}
& Stress & -126.556 & 58.136 & -240.501 & -12.611 & 0.029 \\ \hline
\multirow{2}{*}{ACC z Max} & Calm & 22.816 & 7.245 & 8.616 & 37.016 & 0.002 \\ \cline{2-7}
& Stress & -126.548 & 58.132 & -240.485 & -12.610 & 0.029 \\ \hline
\multirow{2}{*}{ACC L2 Max} & Calm & 22.816 & 7.245 & 8.616 & 37.016 & 0.0016 \\ \cline{2-7}
& Stress & -126.548 & 58.132 & -240.485 & -12.610 & 0.0295 \\ \hline
\multirow{2}{*}{ACC x Kurtosis} & Calm & -3.410 & 1.532 & -6.414 & -0.407 & 0.0261 \\ \cline{2-7}
& Stress & 5.677 & 2.699 & 0.387 & 10.967 & 0.0354 \\ \hline
\multirow{2}{*}{ACC y Kurtosis} & Calm & -3.410 & 1.532 & -6.414 & -0.407 & 0.0261 \\ \cline{2-7}
& Stress & 5.677 & 2.699 & 0.387 & 10.967 & 0.0354 \\ \hline
\multirow{2}{*}{ACC y Permutation Entropy} & Calm & 23.300 & 7.284 & 9.024 & 37.576 & 0.0014 \\ \cline{2-7}
& Stress & -125.477 & 58.184 & -239.515 & -11.439 & 0.0310 \\ \hline
\multirow{2}{*}{ACC y SVD Entropy } & Calm & 23.193 & 7.268 & 8.947 & 37.438 & 0.0014 \\ \cline{2-7}
& Stress & -125.151 & 58.029 & -238.886 & -11.416 & 0.0310 \\ \hline
\multirow{2}{*}{ACC z Interquartile Range} & Calm & 22.815 & 7.246 & 8.614 & 37.016 & 0.0016 \\ \cline{2-7}
& Stress & -126.556 & 58.136 & -240.501 & -12.611 & 0.0295 \\ \hline
\multirow{2}{*}{ACC z Permutation Entropy} & Calm & 23.300 & 7.284 & 9.024 & 37.576 & 0.0014 \\ \cline{2-7}
& Stress & -125.477 & 58.184 & -239.515 & -11.439 & 0.0310 \\ \hline
\multirow{2}{*}{ACC z SVD Entropy} & Calm & 23.193 & 7.268 & 8.947 & 37.438 & 0.0014 \\ \cline{2-7}
& Stress & -125.151 & 58.029 & -238.886 & -11.416 & 0.0310 \\ \hline
\multirow{2}{*}{HRV MeanNN} & Calm & -0.683 & 0.174 & -1.024 & -0.343 & 0.0001 \\ \cline{2-7}
& Stress & -0.713 & 0.267 & -1.236 & -0.189 & 0.0076 \\ \hline
\multirow{2}{*}{HRV Total Power} & Calm & -0.683 & 0.174 & -1.024 & -0.343 & 0.0001 \\ \cline{2-7}
& Stress & -0.713 & 0.267 & -1.237 & -0.189 & 0.0076 \\ \hline
\multirow{2}{*}{HR Entropy} & Calm & -7.940 & 2.856 & -13.538 & -2.342 & 0.0054 \\ \cline{2-7}
& Stress & -13.697 & 4.713 & -22.935 & -4.459 & 0.0037 \\ \hline
\multirow{2}{*}{HR Kurtosis} & Calm & -1.786 & 0.711 & -3.180 & -0.391 & 0.0121 \\ \cline{2-7}
& Stress & -4.839 & 0.998 & -6.796 & -2.883 & 0.0000 \\ \hline
\multirow{2}{*}{HR 95th Percentile} & Calm & -8.294 & 3.396 & -14.951 & -1.637 & 0.0146 \\ \cline{2-7}
& Stress & -15.792 & 6.380 & -28.297 & -3.287 & 0.0133 \\ \hline
\multirow{2}{*}{HR peaks} & Calm & -5.854 & 1.826 & -9.433 & -2.275 & 0.0013 \\ \cline{2-7}
& Stress & -9.513 & 2.077 & -13.585 & -5.441 & 0.0000 \\ \hline
\multirow{2}{*}{HR root mean square} & Calm & 6.143 & 2.507 & 1.230 & 11.056 & 0.0143 \\ \cline{2-7}
& Stress & 11.723 & 4.636 & 2.636 & 20.810 & 0.0115 \\ \hline
\multirow{2}{*}{HR L2 Entropy} & Calm & -4.707 & 2.235 & -9.087 & -0.327 & 0.0352 \\ \cline{2-7}
& Stress & -9.252 & 4.312 & -17.703 & -0.801 & 0.0319 \\ \hline
\multirow{2}{*}{TEMP Root Mean Square} & Calm & 6.814 & 2.387 & 2.135 & 11.492 & 0.0043 \\ \cline{2-7}
& Stress & 14.271 & 4.746 & 4.970 & 23.573 & 0.0026 \\ \hline
\end{tabular}}
\caption{LMM results for common significant features across calm and stress task (WESAD vs Control)}
\label{tab:lmm_detailed_control_wesad}
\end{table}

\subsection{WESAD vs AffectiveRoad}
\begin{table}[h!]
\centering
\scriptsize
\resizebox{0.6\textwidth}{!}{%
\begin{tabular}{|c|c|c|c|c|c|c|}
\hline
\textbf{Feature} & \textbf{Condition} & \textbf{Coef.} & \textbf{Std. Err.} & \textbf{95\% CI Low} & \textbf{95\% CI High} & \textbf{p-value} \\
\hline
\multirow{2}{*}{HRV pNN20} & Calm & 2.178 & 1.106 & 0.010 & 4.346 & 0.049 \\ \cline{2-7}
& Stress & 5.200 & 2.598 & 0.108 & 10.293 & 0.045 \\ \hline
\multirow{2}{*}{BVP L2 Interquartile Range} & Calm & -6.123 & 2.068 & -10.177 & -2.070 & 0.0031 \\ \cline{2-7}
& Stress & -5.459 & 2.675 & -10.702 & -0.217 & 0.0412 \\ \hline
\multirow{2}{*}{BVP L2 Kurtosis} & Calm & 4.478 & 2.062 & 0.437 & 8.518 & 0.0299 \\ \cline{2-7}
& Stress & 9.369 & 4.054 & 1.423 & 17.314 & 0.0208 \\ \hline
\multirow{2}{*}{HR Kurtosis} & Calm & 2.976 & 1.421 & 0.192 & 5.761 & 0.0362 \\ \cline{2-7}
& Stress & 5.600 & 1.659 & 2.348 & 8.851 & 0.0007 \\ \hline
\multirow{2}{*}{HR L2 Peak-to-Peak} & Calm & 4.391 & 2.179 & 0.121 & 8.662 & 0.0439 \\ \cline{2-7}
& Stress & 31.148 & 7.626 & 16.202 & 46.095 & 0.0000 \\ \hline
\multirow{2}{*}{Skin Temp Root Mean Square} & Calm & -13.659 & 4.895 & -23.254 & -4.064 & 0.0053 \\ \cline{2-7}
& Stress & -15.002 & 5.610 & -25.998 & -4.006 & 0.0075 \\ \hline
\multirow{2}{*}{Tonic Peaks} & Calm & 2.178 & 1.106 & 0.010 & 4.346 & 0.0490 \\ \cline{2-7}
& Stress & 5.200 & 2.598 & 0.108 & 10.293 & 0.0453 \\ \hline
\end{tabular}}
\caption{LMM results for common significant features across stress and calm tasks (WESAD vs AffectiveRoad)}
\label{tab:lmm_detailed_wesad_affectiveroad}
\end{table}

\end{document}